\documentclass[conference]{IEEEtran}
\IEEEoverridecommandlockouts
\usepackage{cite}
\usepackage{amsmath,amssymb,amsfonts}
\usepackage{graphicx}
\usepackage{textcomp}
\usepackage{xcolor}
\usepackage{float}
\usepackage{subfig}
\usepackage{booktabs}
\usepackage{hyperref} 
\usepackage{algorithm}
\usepackage{algorithmic}
\usepackage{url}
\usepackage{colortbl} 
\def\BibTeX{{\rm B\kern-.05em{\sc i\kern-.025em b}\kern-.08em
    T\kern-.1667em\lower.7ex\hbox{E}\kern-.125emX}}
\begin{document}

\title{GraphPrompter: Multi-stage Adaptive Prompt Optimization for Graph In-Context Learning}

\author {
    \IEEEauthorblockN{
    Rui Lv\textsuperscript{\rm 1},
    Zaixi Zhang\textsuperscript{\rm 1},
    Kai Zhang\textsuperscript{\rm 1},
    Qi Liu\textsuperscript{\rm 1}*\thanks{*Corresponding author},
    Weibo Gao\textsuperscript{\rm 1},\\
    Jiawei Liu\textsuperscript{\rm 1},
    Jiaxia Yan\textsuperscript{\rm 1},
    Linan Yue\textsuperscript{\rm 1},
    Fangzhou Yao\textsuperscript{\rm 1}}
    \IEEEauthorblockA{
    \textsuperscript{\rm 1}State Key Laboratory of Cognitive Intelligence, University of Science and Technology of China\\
    \{lvrui2018, zaixi\}@mail.ustc.edu.cn, kkzhang08@ustc.edu.cn, qiliuql@ustc.edu.cn,\\
    \{weibogao, ljw1222, jiaxianyan, lnyue, fangzhouyao\}@mail.ustc.edu.cn
    }
}
\maketitle

\begin{abstract}
Graph In-Context Learning, with the ability to adapt pre-trained graph models to novel and diverse downstream graphs without updating any parameters, has gained much attention in the community. The key to graph in-context learning is to perform downstream graphs conditioned on chosen prompt examples.
Existing methods randomly select subgraphs or edges as prompts, leading to noisy graph prompts and inferior model performance. 
Additionally, due to the gap between pre-training and testing graphs, when the number of classes in the testing graphs is much greater than that in the training, the in-context learning ability will also significantly deteriorate.
To tackle the aforementioned challenges, we develop a multi-stage adaptive prompt optimization method GraphPrompter, which optimizes the entire process of generating, selecting, and using graph prompts for better in-context learning capabilities. 
Firstly, Prompt Generator introduces a reconstruction layer to highlight the most informative edges and reduce irrelevant noise for graph prompt construction. Furthermore, in the selection stage, Prompt Selector employs the $k$-nearest neighbors algorithm and pre-trained selection layers to dynamically choose appropriate samples and minimize the influence of irrelevant prompts. Finally, we leverage a Prompt Augmenter with a cache replacement strategy to enhance the generalization capability of the pre-trained model on new datasets. 
Extensive experiments show that GraphPrompter effectively enhances the in-context learning ability of graph models. On average across all the settings, our approach surpasses the state-of-the-art baselines by over 8\%. Our code is released at \href{https://github.com/karin0018/GraphPrompter}{https://github.com/karin0018/GraphPrompter}.
  
\end{abstract}


\section{Introduction}
\label{intro}
One of the most fascinating properties of Large Language Models (LLMs) is its In-Context Learning capability~\cite{brown2020language,dong2022survey}. It refers to the ability of a pre-trained LLM to achieve competitive results on downstream tasks given only a few prompt examples during the prediction phase, without updating the model weights through fine-tuning approaches. Recently, there have been efforts to transfer this In-Context learning capability from large language models to graph models~\cite{huang2023Prodigy, sun2023all, oneforall}. Out of these methods, Prodigy~\cite{huang2023Prodigy} and One For All (OFA)~\cite{oneforall} stand out as the most effective frameworks that unify diverse levels of graph-related tasks and achieve competitive in-context learning performance. 
\begin{figure}

  \centering
  \includegraphics[width=\linewidth]{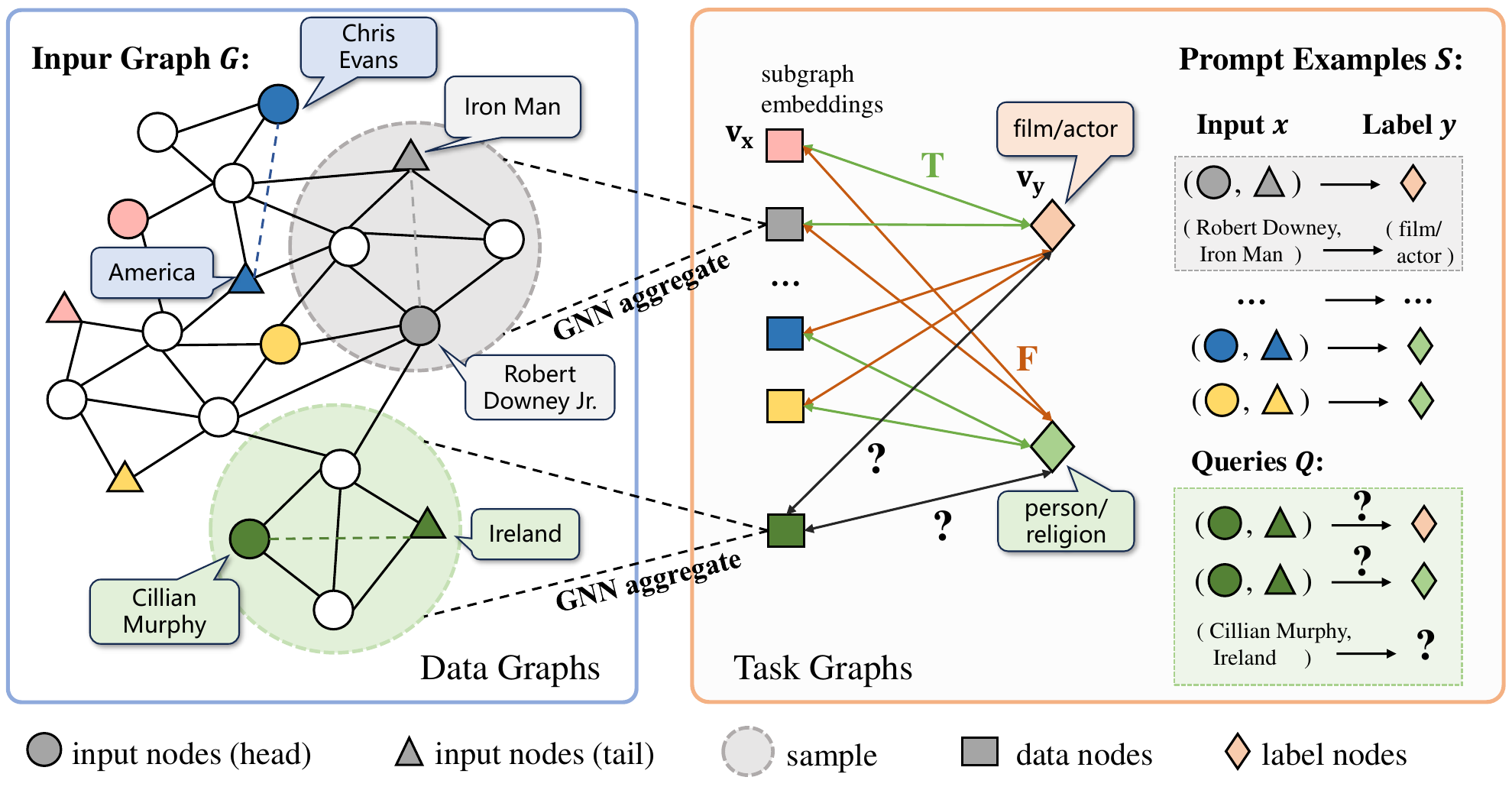}

  \caption{Graph In-Context Learning (edge classification as an example) with random prompts selection. The graph prompts are sampled $l$-hop subgraphs (left) and the downstream tasks are reformulated as the edge label predictions between the query data graph and label nodes (right).}
  \label{fig:Prodigy}

\end{figure}
Generally, the graph in-context learning architecture can be divided into two main parts including data/prompt graph construction and task graph prediction (see Figure~\ref{fig:Prodigy} as an example for edge classification). For the given source graph $G$, randomly selected head/tail nodes consist of the input prompt examples $S$  and Queries $Q$.  After constructing context subgraph embeddings for each pair of nodes by \emph{Data Graph}, these data nodes $v_x$ will be connected with their label nodes $v_y$ in the \emph{Task Graph}. 
The downstream tasks (e.g., node/graph classifications and link predictions) are unified as predicting the edge labels connecting the label nodes and the query data graph or comparing the prediction values of the label nodes. {For example, sample "(Robert Downdey Jr., Iron Man) $\rightarrow$ film/actor"  and "(Chris Evans, America) $\rightarrow$ person/religion" as prompts, the query is to get the relationship (edge label) between "Cillian Murphy" and "Ireland".}

{However, the limitations of prompt generation and selection strategies of existing graph in-context learning works severely restrict their generalization ability. For instance, in the prompt generation stage, Prodigy \cite{huang2023Prodigy} uses the random walk algorithm to construct the prompt subgraph, which does not consider to optimal subgraph structure. However, there is task-irrelevant information mixed into nodes’ neighborhoods, making learned models suffer from sub-optimal generalization performance~\cite{hier-graph-transformer, zheng2020robust,zha2023data,luo2021learning,jin2020graph}. In the prompt selection stage, as shown in Figure~\ref{fig:Prodigy}, when selecting prompts from the candidate set for each query, existing methods also randomly select $k$ examples. Such a method does not adapt to the query's unique characteristics, which further limits the graph in-context learning ability. Furthermore, in the context of few-shot scenarios, due to the limited number of labeled samples and the gap between pre-training and testing graphs, when the number of classes in the testing graphs is much greater than that in the training, the in-context learning ability of the pre-trained model will also significantly decrease.}
 
{The prompt optimization strategy has been extensively studied in LLMs and proved to be effective in improving the generalization ability of the language models. However, there is a lack of prompt optimization strategies on graph models. In the task of domain generalization of graphs, the pre-trained and tested graph data structures are completely different, and it is difficult for a graph model to be able to map both upstream and downstream data samples into the same hidden space as LLMs. This limits the ability of graph models to in-context learning.}
{To tackle the above challenges, we develop a novel multi-stage prompt optimization method \textbf{GraphPrompter}, to adaptively construct the most appropriate in-context prompt graphs for each query.}

\textcolor{black}{Specifically, we have designed corresponding modules to improve the model during the generation, selection, and use stages of the prompt, named \textbf{Prompt Generator}, \textbf{Prompt Selector}, and \textbf{Prompt Augmenter}. First, in the prompt generation stage, the Prompt Generator utilizes the pre-trained graph reconstruction layers to learn the weights of the edges of the subgraphs obtained by sampling, reconstructing the weights of the neighboring nodes to the target nodes, thereby filtering some task-irrelevant nodes and edges, enhancing the quality of the generated prompt.} 

The main challenge for the Prompt Selector is how to judge the suitable examples for queries (nodes/edges) in graph models. {We first design the selection layers to learn the importance of each prompt for the category center during pre-training. However, since the downstream graphs is quite different with the pre-training graph, the pre-trained selection layers limit the models generalization on these test graphs. Inspired by retrieval augmentation works in LLMs~\cite{liu-etal-2022-makes, zhao2021calibrate,rubin-etal-2022-learning,su2023selective,wu-etal-2023-self}, which using semantic similarity of sentences as a criterion to evaluate the quality of prompts and making pre-trained models adapt to different downstream domain effectively, we use $k$-Nearest Neighbor ($k$NN) algorithm to calculate the similarity of each prompt for the query and choose the top-$k$ similar prompt graph. This method can be used effectively and doesn't need to update any parameters in inference. Finally, we combine the selection layers output and $k$NN algorithm to adaptive filter the appropriate prompts, making the prompt selection process more accurate and efficient.}

{When dealing with a large number of categories in the test dataset, using fixed candidate prompts can limit the model's adaptive ability. So, we aim to obtain a dynamic candidate prompts set to enhance the generalization ability of the pre-trained model on new datasets. To achieve that, we propose an online prompts augmentation method in Prompt Augmenter.} Specifically, we incorporate test samples with predicted labels (pseudo-labels) into the prompt set. To control the quantity of task graph inputs, we used a cache to store these online samples and designed the least frequently used (LFU) replacement strategy to update it. Despite its simplicity, this method can leverage knowledge about the target domain by using off-the-shelf test-time data and improve performance. Our method has been shown to effectively enhance the in-context learning ability of graph models through empirical evidence. On average across all setups, our approach surpasses the baseline in-context learning accuracy by 8\%. 

Overall, our contributions are summarized below:
\begin{itemize}
    \item To the best of our knowledge, we are the first research to explore the concept of prompt optimization on graph models and propose a novel framework in this area that combines offline and online strategies, which optimizes the entire process of generating, selecting, and using graph prompts for better graph in-context learning capabilities.

    \item Our experimental results demonstrate that we are able to improve performance by an average of 8\% on all datasets compared to baseline. 
    \item We provide technical and performance insights on the use of prompt optimization in the graph domain, which is instructive for research and practice in related areas.
\end{itemize}

\section{Related Work}
{\subsection{Graph Pretraining} Most existing Graph Neural Networks (GNNs) \cite{ graph-self-superLearning, gat2018,kipf2017semi,E2GCL,GraphRARE,zhou2024fast} adopt the message-passing framework and use permutation-invariant local aggregation schemes to update node representations. For example, Graph Convolutional Networks (GCNs) \cite{kipf2017semi} average features from neighboring nodes, while Graph Attention Networks (GATs) \cite{gat2018} use an attention mechanism to assign different weights to neighbors. GraphSAGE \cite{hamilton2017inductive} samples a fixed-size set of neighbors and aggregates their features, allowing for faster and more scalable GNN training. \cite{E2GCL, zhou2024fast} focuses on improving the inference speed of GNNs, while \cite{GraphRARE} introduces a novel approach that applies reinforcement learning to denoise graph structures. However, these models often experience a significant performance drop in domain adaptation settings.}
Recent work in graph pretraining has aimed at improving this issue by learning better encoders that can perform specific pretraining tasks. Most of these methods~\cite{Hu*2020Strategies, you2020graph, hu2020gpt, qiu2020gcc, graphaug2024} focus on tasks such as masked feature prediction~\cite{Hu*2020Strategies} and paired graph classification~\cite{you2020graph}. To adapt these pre-trained models to different downstream graphs, a large amount of task-specific data is typically required to fine-tune a classification head on top of the encoder.
In contrast, Prodigy~\cite{huang2023Prodigy} has been introduced as a pretraining approach that aims to equip the model with general in-context learning capabilities. This allows the pre-trained model to be used directly for various downstream graphs without the need for gradient-based updates. However, the generalization ability of Prodigy is limited by its prompt selection strategy, leaving room for further improvement.

\subsection{Graph prompt learning}  
Graph In-Context Learning is a kind of graph prompt learning. Inspired by prompt learning in natural language processing (NLP), which effectively utilizes prior knowledge for various NLP tasks, researchers in the graph domain have recently explored graph prompt learning \cite{zhu2023graphcontrol, cooperative-classification, fang2023universal, sun2023graph,allinone,huang2023Prodigy,sun2022gppt,graph-prompt,gong2023prompt,fewshotcf23,ge2023enhancing,ma2023hetgpt}.
It reformulates downstream graph tasks into those solved during the pre-training phase using graph prompts and bridges the gap between pre-trained models and different graph tasks. Existing graph prompt learning approaches can be divided into two types based on prompt design~\cite{sun2023graph}: Prompt Token methods and Prompt Graph methods. The former method treats the graph prompt as additional features added to the original graph features, like GraphPrompt~\cite{graph-prompt} and All-in-One (ProG)~\cite{allinone}. 
These methods focus on cross-task research and achieve effective adaptation to downstream tasks through prompt fine-tuning. However, they can hardly work in challenging cross-domain tasks, where the downstream task dataset and the pre-training dataset differ completely. It becomes particularly evident in scenarios with limited labeled samples. 

Compared to Prompt Token methods that treat prompts as learnable vectors, the Prompt Graph methods employ subgraphs surrounding the target nodes as prompts, which allows for preserving the structural information of the graph within the prompt (e.g., Prodigy~\cite{huang2023Prodigy} and One-For-All (OFA)~\cite{oneforall}). It also facilitates the efficient transfer of the in-context learning paradigm from LLMs to the graph domain. Particularly in low-resource scenarios, these approaches can achieve impressive results without the need for fine-tuning any parameters~\cite{huang2023Prodigy}. This is something that Prompt Token methods cannot achieve~\cite{sun2022gppt,graph-prompt,allinone}.
However, existing methods commonly employ simple approaches for subgraph selection and construction, such as random sampling and prompt selection. In this paper, we propose a novel method called GraphPrompter to enhance the construction, selection, and online augmentation of the prompt graph, which aims to improve the overall performance of graph models in in-context learning. 
\subsection{In-context Example Selection of Large Language Models.} 
In-context learning with large language models has recently received an increasing amount of interest, partly due to its flexibility and sample efficiency~\cite{liu2023pre,liu-etal-2022-makes}. Several recent works proposed methods to improve in-context learning in many aspects: e.g., meta-training~\cite{min-etal-2022-metaicl,chen-etal-2022-meta} or task formulation~\cite{holtzman2021surface}. In this paradigm, the choice of in-context (i.e., demonstration) examples has been shown crucial~\cite{hashimoto2018retrieve,rubin-etal-2022-learning,liu2023pre,su2023selective,wu-etal-2023-self,GripRank}. Inspired by these research, we propose an adaptive method based on graph topology (GraphPrompter) to construct appropriate prompts for every query.
\subsection{Test-time adaptation.} 
Test-time adaptive methods are recently proposed to utilize target samples, which adjust model parameters based on unsupervised objectives such as entropy minimization~\cite{wang2021tent,zhang2021memo} or update a prototype for each class~\cite{iwasawa2021testtime}. These methods are easily integrated with classification algorithms and can utilize information from the target dataset by incorporating off-the-shelf test-time data, resulting in performance improvement. Building upon these inspiring studies, we have introduced an online prompts augmentation approach that leverages a cache replacement strategy to augment the prompt sets and enhance the generalization capacity of pre-training graph models.

\section{Preliminaries}
\begin{figure*}[tb]
  \centering
  \includegraphics[width=\linewidth]{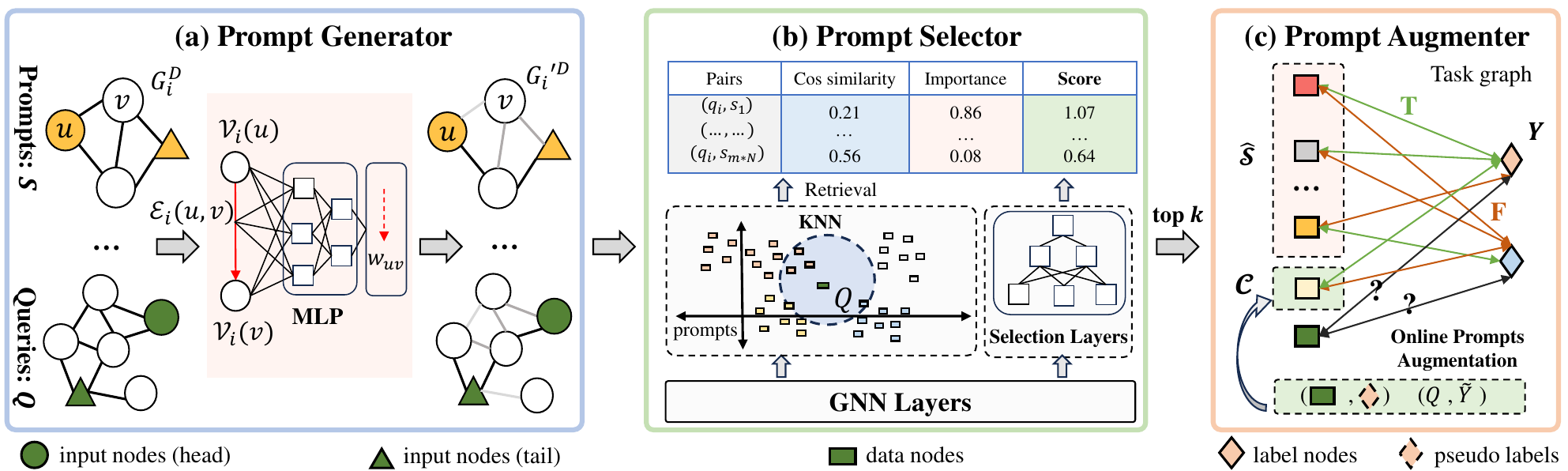}
  \caption{{An overview of the GraphPrompter method. Overall the method can be divided into three components: (a) \textit{Prompt Generator}. We select candidate prompt subgraphs $\mathcal{S}$ and query subgraphs $\mathcal{Q}$ and filter their edges by the pre-training edge weights(represented by varying shades). (b) In \textit{Prompt Selector} stage, after obtaining embeddings for subgraphs through GNN layers, we computed the probability (Score) of each prompt being selected as the combination of the $k$NN similarity and the importance value obtained from the pre-trained selection layer to adaptively select the top-$k$ suitable prompts. These prompts form the new prompt set $\mathcal{\hat{S}}$. (c) In \textit{Prompt Augmenter}, we utilize a cache $\mathcal{C}$ to store online test samples and their pseudo labels, which helps enhance the prompt set dynamically. We predict the query using adaptive prompt selection and optionally consider prompts from the cache. All the model parameters are learned in the pretraining phase.}}
  \label{fig:method}
\end{figure*}

\begin{table}[htbp]
\centering
\begin{tabular}{|l|l|}
\hline
\textbf{Notation} & \textbf{Description} \\ \hline
$\mathcal{V, E, R}$ & The set of nodes, edges, and relations \\ \hline
$e=(u,r,v)\in {E}$ & An edge consists of $u \in \mathcal{V}$, $r \in \mathcal{R}$, $v \in \mathcal{V}$ \\ \hline
$\mathcal{G=(V, E, R)}$ & The input graph \\ \hline
$\mathcal{X}$ & The input subgraph set \\ \hline
$x_i = (\mathcal{V}_i,\mathcal{E}_i,\mathcal{R}_i) \in \mathcal{X}$ & An input node/edge \\ \hline
${G}_i^D$ & Data graph: $l$-hop neighborhood of input $x_i$ \\ \hline
${G}^T$ & Task graph \\ \hline
$\mathcal{Y}$ & A set of classes \\ \hline
$\mathcal{S}$ & Prompt set \\ \hline
$\mathcal{C}$ & Cache \\ \hline
$\mathcal{Q}$ & Query set \\ \hline
$N$ & The size of candidate prompt set \\ \hline
$k$ & The number of used prompts ($k$-shot) \\ \hline
$n$ & Query number \\ \hline
$l$ & The distance of neighbors, i.e., $l$-hop \\ \hline
$m$ & Classes number \\ \hline
$p$ & One of graph prompt in $\mathcal{S}$ \\ \hline
$q$ & One of query in $\mathcal{Q}$ \\ \hline
\end{tabular}

\caption{Frequently-used notations.}
\end{table}

In this study, our primary emphasis is on leveraging graph in-context learning (GICL) to tackle node and edge classification problems, employing the proposed adaptive prompting techniques (GraphPrompter).
In this section, we first introduce the concrete definition of graph and GICL. Then, we further describe the formulations of data graphs and task graphs.
\subsection{Task Definition}
The tasks of the GICL framework are defined as follows:\\
\textbf{Definition 1: Graph.} A graph can be defined as $\mathcal{G=(V, E, R)}$, where $\mathcal{V, E, R}$ represents the set of nodes, edges, and relations. An edge $e=(u,r,v)\in \mathcal{E}$ consist of a subject $u \in \mathcal{V}$, a relation $r \in \mathcal{R}$, and an object $v \in \mathcal{V}$. 
\\
\textbf{Definition 2: Graph In-Context Learning.} Given a set of classes $\mathcal{Y}$, a normal classification task is to predict the label $y \in \mathcal{Y}$ of each input $x_i \in {X}$, where $x_i = (\mathcal{V}_i,\mathcal{E}_i,\mathcal{R}_i)$. For example, for the node classification task, ${x}_i$ only consists of the input node that we aim to make predictions on, i.e., $|\mathcal{V}_i|=1$, $|\mathcal{E}_i|=0$ and $|\mathcal{R}_i|=0$; for edge classification, it consists of node pair (head, tail), i.e., $|\mathcal{V}_i|=2$, $|\mathcal{E}_i|=1$ and $|\mathcal{R}_i|=1$. For the GICL framework, the classification tasks over graphs are defined with few-shot prompting. Such as a $k$-shot prompt with a downstream $m$-way classification, $|\mathcal{Y}|=m$, we use a small number of input-label pairs $\mathcal{S}=\{(x_i,y_i)\}^{m\cdot k}_{i=1}$ as the prompt examples, and the queries is a set of nodes $\mathcal{Q}=\{x_i\}_{i=1}^n$ that we want to predict labels for. Both the prompts $\mathcal{S}$ and queries $\mathcal{Q}$ are the inputs given to the graph model.

\subsection{Prompting graph representation}
As we introduced in Section~\ref{intro}, a $k$-shot prompt over graphs for a $m$-way classification task is composed of data graphs and task graphs. The two graphs are defined as follows: \\
\textbf{Data Graphs.} We defined the data graphs are the contextualization of each datapoint $x_i= (\mathcal{V}_i, \mathcal{E}_i, \mathcal{R}_i)$ in inputs by sampling $l$-hop neighborhood of $\mathcal{V}$ in the source graph $\mathcal{G}$, which can be represented as ${G}_i^D = (\mathcal{V}_i^D,\mathcal{E}_i^D,\mathcal{R}_i^D) \sim \oplus_{i=0}^{l} \text{Neighbor}(\mathcal{V}_i,\mathcal{G},i)$, where $\mathcal{V}_i \subseteq \mathcal{V}_i^D \subseteq \mathcal{V}, \mathcal{E}_i \subseteq \mathcal{E}_i^D \subseteq \mathcal{E}, \mathcal{R}_i \subseteq \mathcal{R}_i^D \subseteq \mathcal{R}$, and Neighbor is a function that returns the exact $i$-hop neighbors of each node in $\mathcal{V}_i$. Then, we use reconstruction layers to learn the edge weight $\mathcal{W}_i^D = {w_{uv},\forall e_{uv} \in \mathcal{E}_i^D}$, the final data graphs is represented as $G{'}_i^D=(\mathcal{V}_i^D,\mathcal{E}_i^D,\mathcal{R}_i^D)$. \\
\textbf{Task Graphs.} After achieving contextual data graphs for each input $x_i$, we construct task graph ${G}^T$ to capture the connection among the inputs and the labels. For each data graph ${G}{'}_i^D$, 
we use $G_i$ to represent the embedding after aggregating by GNNs. 
Overall, a task graph contains $m\cdot k + n$ data nodes ($m\cdot k$ prompts and $n$ queries) and $m$ label nodes, as shown in Figure~\ref{fig:Prodigy}. Specifically, task graph is a bipartite graph composed of data nodes and label nodes. Each edge has two attributes: one distinguishes query from prompts, the other indicates the class. For example, each prompt node connects to all label nodes, with edge attributes set to "T" for the true label and "F" otherwise.

\section{Methodology}
\label{method}

{In this section, we present our proposed GraphPrompter, the multi-stage adaptive graph prompt optimization strategy to enhance the in-context learning ability over graph models. Our method can be divided into three stages, including \textbf{Prompt Generator} (Section \ref{model-arc}), \textbf{Prompt Selector} (Section \ref{prompt_selector}), and \textbf{Prompt Augmenter} (Section \ref{online-aug}). In the first stage, 
we random sample a set of candidate prompts, denoted as $\mathcal{S}$, from the source graph $\mathcal{G}$. To generate subgraph embeddings with higher equality, we employ reconstruction layers to learn new edge weights. Then, we leverage GNN layers to obtain the embeddings of the reconstructed subgraph.  In the second stage, we aim to construct an improved prompt set, denoted as $\hat{S}$. Specifically, we design the pre-trained selection layers for learning the importance of each prompt node, and we introduce an adaptive $k$-nearest neighbor approach to retrieve the most appropriate $k$ prompts to enhance the generalization ability in unseen downstream graphs. In the last stage, we proposed an online strategy to further dynamically enhance the prompt set in the task graph. Specifically, we make use of a cache, represented as $\mathcal{C}$, to store online test samples along with their pseudo labels and use the least frequently used (LFU) replacement algorithm to update it. This multi-stage graph prompt optimization strategy allows us to effectively utilize relevant prompts and leverage the cache to augment the prompt set, resulting in more accurate and contextually appropriate predictions. Figure~\ref{fig:method} illustrates the overall process of our proposed method.}

\subsection{Prompt Generator}
\label{model-arc}
\subsubsection{\textbf{Prompt graph generation.}} We first randomly select $N$ data points as the candidate prompts nodes and $n$ data points as the queries. The data graphs are contextualization by using the random walk algorithm to sample $l$-hop neighborhood in source graph $\mathcal{G}$. {The random walk algorithm starts from the selected node, adds its neighboring nodes to the subgraph. Then, randomly chooses a direction to move to the next node. The neighbors of this node are added to the subgraph, with duplicates removed. This process is repeated \(l\) times, and the algorithm terminates if the number of nodes in the subgraph reaches the preset limit.}
For each datapoint $x_i$, the sampled subgraph ${G}_i^D$ can be represented as:
\begin{equation}
    {G}_i^D = (\mathcal{V}_i^D,\mathcal{E}_i^D,\mathcal{R}_i^D) \sim \oplus_{i=0}^{l} \text{Neighbor}(\mathcal{V}_i,\mathcal{G},i).
\label{gid}
\end{equation}
\subsubsection{\textbf{Prompt graph reconstruction.}} 
Instead of directly using the sampled $l$-hop subgraphs, like previous works (e.g., Prodigy~\cite{huang2023Prodigy} and OFA~\cite{oneforall}), we reconstruct the weights of the edges in each subgraph to adaptively extract the most useful topological information and filter the noise. The edge weights are assigned based on the node/edge embedding and we jointly train the reweighting modules along with the graph model. Specifically, the reconstruction method can be split into the following 3 steps:\\
\textbf{(1)} Let $\mathcal{V}_i(u), u\in \mathcal{V}_i^D $ be the node $u$ embedding of the subgraph $G_i^D$, $\mathcal{E}_i(u,v), v \in \text{Neighbor}(u)$ is the edge $(u,v)$ embedding, 
\begin{equation}
    z_{uv} = \text{MLP}_{\phi}(\mathcal{V}_i(u), \mathcal{V}_i(v), \mathcal{E}_i(u,v)),
\label{zuv}
\end{equation}
where MLP is a multi-layer neural network with parameters $\phi$. Specifically, for the node classification task, we use the concat of $\mathcal{V}_i(u)$ and $\mathcal{V}_i(v)$ as the input of MLP, for the edge classification task, each edge have its specific initial embedding, so we use the $\mathcal{E}_i(u,v)$ as the input.\\
\textbf{(2)} $\forall v \in \text{Neighbor}(u)$, we use a sigmoid function to compute the weights of the edge $w_{uv}$:
\begin{equation}
\label{wuv}
    w_{uv} = \frac{1}{1+e^{-z_{uv}}}.
\end{equation}
\textbf{(3)} Finally, we achieve a single subgraph embedding $G_i$ by using a GNN to aggregate the data graph with edge weights $G{'}_i^D = (\mathcal{V}_i^D,\mathcal{E}_i^D, \mathcal{R}_i^D, \mathcal{W}_i^D)$, where $\mathcal{W}_i^D = {w_{uv},\forall e_{uv} \in \mathcal{E}_i^D}$:
\begin{equation}
\label{gi}
    G_i = \text{GNN}_D(G{'}_i^D), 
\end{equation}
where $\text{GNN}_D$ denote a message passing GNN module.\\
{$G_i$ used in the subgraph embeddings in the subsequent module, and the parameters in $\text{GNN}_D$ are trained by the total optimization object.}

\subsection{Prompt Selector}
\label{prompt_selector}
\subsubsection{\textbf{Pre-training selection layers.}}
In the generation paradigm of in-context learning, the quality of prompts has a significant impact on the final prediction results. This has been validated in related research on large language models~\cite{hashimoto2018retrieve,su2023selective,wu-etal-2023-self}. This intuition also holds in the network structure of Prodigy, because the label embeddings in the task graph are aggregated from prompts, and the final prediction of the query is determined by computing the similarity with label embeddings. Therefore, in the graph model, the selection of prompts also affects the performance of the pre-trained model on downstream graphs. However, Prodigy's approach of randomly selecting prompts ignores the contextual information of the query, thereby limiting the model's performance.
Instead of randomly selecting prompts from the candidate prompt set, we decided to use learnable layers to choose the most relevant prompt graph. Specifically, for each candidate subgraph embedding $G_{p} \in G_S$, where $G_S$ noted the all candidate subgraph embeddings, $1\leq p \leq N$. We get their importance from the selection layers:
\begin{equation}
\label{ip}
    I_{p} = \sigma(\text{MLP}_{\theta}(G_{p})),
\end{equation}
where $\sigma$ is a sigmoid function and MLP is a multi-layer neural network with parameters $\theta$. Then, we use the importance value to get the weighted data node embeddings as the input of the task graph, note as $G_{S_I} = \{G_{p}' |G_{p}' = G_{p} * I_{p}, \forall G_{p}\in G_S\}$. However, the selection layers are not updated in the test phase, limiting its generalization performance on unseen downstream graphs, so we need an offline algorithm in the inference phase to improve it.
\subsubsection{\textbf{$k$-nearest neighbor retrieval.}}
In order to enhance the generalization performance of large-scale pre-training models on test graphs, we leverage retrieval-enhanced techniques to dynamically determine the most appropriate in-context prompt examples for each query during the inference stage. 
As we introduced in Section~\ref{model-arc}, we can achieve subgraph embeddings for both prompts and queries after data graph neural network ${GNN}_D$. At this step, we proposed to utilize the $k$-nearest neighbors algorithm to compute the similarity of embeddings between the query and prompt, using it as one of the criteria for selecting prompts.
Specifically, as illustrated in Figure~\ref{fig:method}, for each query $q \in Q$, we compute the similarity between each prompt $G_{p} \in G_{S_I}$:
\begin{equation}
\label{simpq}
    \text{sim}(p,q) = \text{Cos}(G_{p},G_{q}),
\end{equation}
where $G_{p}$ is the subgraph embedding of the $p$th candidate prompts, $G_{q}$ is the subgraph embedding of $q$th query. Note that we use cosine similarity to calculate the similarities, which can be substituted by other distance metrics, like Euclidean distance or Manhattan distance.

Then, we combine the pre-trained selection layers output and $k$NN retrieval results to filter the most appropriate prompts. Specifically, for each query $q$, the probabilities of the candidate prompts $p$ is:
\begin{equation}
\label{scorepq}
    \text{score}(p,q) = \text{sim}(p,q) + I_{p} * I_{q},
\end{equation}
Furthermore, we take a voting mechanism to consider the overall preferred prompts for all the queries from the set $\mathcal{Q}$, where $|\mathcal{Q}|=n$.
Specifically, each query casts votes for the $k$ importance prompts to it, with the number of votes determined by the corresponding $\text{score}(p, q)$ value:
\begin{equation}
\label{votesp}
    \text{Votes}(p) = \sum\limits_{q \in \mathcal{Q}} \mathbf{1}_{[p \in \mathcal{T}(q)]} \text{score}(p,q),
\end{equation}
where $\mathcal{T}(q) = \text{top-}k\text{-prompts}\{\text{score}(p_i,q), \forall p_i \in \mathcal{S}\}$. Finally, we select the top $k$ prompts with the highest number of votes to form the set $\mathcal{\hat{S}}$, which along with the embeddings of all queries, serves as the input to the next component.

\subsection{Prompt Augmenter}
\label{online-aug}
As mentioned in Section~\ref{intro}, the model's in-context learning ability greatly depends on the quality and diversity of the graph dataset used for pre-training. To ensure robust performance, the pre-training dataset should be extensive and encompass a wide range of graph structures. Nevertheless, due to limitations in available datasets and models, the model may not encounter all possible graph structures during pre-training. Consequently, the model's heavy reliance on the pre-training data results in weaker generalization when applied to new graph datasets. For instance, Prodigy achieved over 88\% accuracy on prediction tasks with 5 labels in new datasets (such as FB15K-237~\cite{xiong-etal-2018-one} and NELL~\cite{xiong-etal-2018-one}) when it was pre-trained on a dataset with only 15 labeled categories for edge classification. However, when the prediction task involved more than 15 categories i.e.,40, the accuracy dropped to just 60\%. 

Inspired by research of test-time adaptation \cite{iwasawa2021testtime, wang2021tent, dubey2021adaptive, AdaNPC}, we propose a novel online prompting sample augmentation based on \textit{cache replacement strategy} \cite{LFU} to mitigate this issue.
Specifically, as shown in Figure~\ref{fig:method}, we aim to incorporate test samples with predicted labels (pseudo-labels) $\hat{\mathcal{Y}}$ into the prompts set $\hat{\mathcal{S}}$.
{To manage the quantity of task graph inputs, we have devised a cache ${C}$ of size $c$ to store the online test sample embeddings $\hat{\mathcal{Q}}$ along with their pseudo-labels $\hat{\mathcal{Y}}$, where $\hat{\mathcal{Q}}$ consists of the query sample in $\mathcal{Q}$ which have the most confidence probability in its predicted labels, we can achieve $|\hat{{Q}}| \le m$ ($m$ is the number of labels).} Thus, the set of prompts used online prompts augmentation can be represented by:
\begin{equation}
\label{s'}
\hat{\mathcal{S}'} = \hat{\mathcal{S}} \cup \mathcal{C},
\end{equation}
where $|\hat{\mathcal{S}'}| = k+c$, and the details of cache size setting will be discussed in Sec~\ref{sec:parameter}.

In addition, by leveraging the spatial locality of graph data sampling, we employ the LFU (Least Frequently Used) replacement algorithm~\cite{LFU} to dynamically update these online samples. Specifically, we calculate the similarity between $q_j \in \mathcal{Q}$ and the embeddings of prompts in the cache $q_c \in {C}$, and entries with the top-$k$ highest similarity scores are considered hits. We use this information to update the usage frequency of entries in the cache. Although straightforward in its approach, the online prompts augmentation technique possesses the capacity to harness valuable knowledge about the target domain by incorporating readily available test-time data, leading to performance enhancement. By implementing this procedure, the adapted decision boundary steers clear of the densely populated regions within the target domain, consequently diminishing ambiguity (or entropy) in predictions, which is closely linked to reducing classification errors~\cite{wang2021tent}.

Finally, we input the obtained prompt set $S_I$(in pertaining) or $\hat{\mathcal{S}'}$ (in testing) and query set $\mathcal{Q}$ together into the task graph ${G}^T$ to obtain the prediction, where ${G}^T$ enabling the fusion of prompts associated with each class, thereby yielding the corresponding label embedding:
\begin{equation}
\label{h}
    \mathbf{H} = \text{GNN}_T({G}^T(S_I | \hat{\mathcal{S}'},\mathcal{Q})),
\end{equation}
where $\mathbf{H}$ is the embedding matrix for each node in the task graph. Subsequently, a nearest neighbor algorithm is employed to discern the class affiliation of a given query. This entails computing the cosine similarity ($\text{Cos}(\cdot,\cdot)$) between the query node embedding within the task graph and the label embeddings. Ultimately, the label exhibiting the highest value of similarity to the query $\mathbf{h}_{q}$ is selected as the prediction $y_q$:
\begin{equation}
\label{yq}
    y_q = \underset{y \in \mathcal{Y}}{\arg \text{max}} ~ \text{Cos}(\mathbf{h}_{q},\mathbf{h}_y),
\end{equation}

\subsection{Pretrain task}
\begin{algorithm}[t]
\caption{GraphPrompter: Training Stage}
\label{alg:training-stage}
\textbf{Input:} Source Graph $\mathcal{G}$, all training steps \\
\textbf{Output: } Pre-trained parameters
\begin{algorithmic}[1]
\STATE Initialize all parameters;
\FOR{step $= 1, 2, \dots, \text{Steps}$}
    \STATE Sample the subgraphs from source graph $\mathcal{G}$ via Eq~\ref{gid};
    \STATE {Calculating the edge weights of subgraphs via Eq~\ref{zuv},\ref{wuv};
    \STATE Calculating their embeddings via GNN in Eq~\ref{gi};
    \STATE Using selection layers to obtain the importance of prompts via Eq~\ref{ip};
    \STATE Construct Task graph;
    \STATE Joint optimization of GraphPrompter to minimize task loss via Eq~\ref{l};}
\ENDFOR
\RETURN All parameters;
\end{algorithmic}
\end{algorithm}
Following \cite{huang2023Prodigy}, we use two pretraining tasks, namely \textit{Neighbor Matching} and \textit{Multi-Task} are leveraged in our method. The primary objective of the \textit{Neighbor Matching} task is to classify the local neighborhood to which a given node belongs. Each local neighborhood is defined by the set of example nodes that are associated with that particular neighbor. The query set is defined as $\mathcal{Q}_{NM}$ and the loss function employed for the \textit{Neighbor Matching} task is:
\begin{equation}
\label{lnm}
    {L}^{NM} = \underset{q \in \mathcal{Q}_{NM}}{\mathbb{E}} \text{CrossEntropy}(P_{NM_q},y_{NM_q}),
\end{equation}
where $P_{NM_q}$ is the probability of predicted by the prompt graph, $y_{NM_q}$ is the corresponding label of query $q$. 

\textit{Multi-Task} focuses on constructing supervised tasks in the form of few-shot prompts along with their corresponding labels. The objective is to enable the model to learn from limited examples and generalize to new instances. The query set is defined as $Q_{MT}$ and the loss function utilized for the \textit{Multi-Task} is as follows:
\begin{equation}
\label{lmt}
    \mathcal{L}^{MT} = \underset{q \in \mathcal{Q}_{MT}}{\mathbb{E}} \text{CrossEntropy}(P_{MT_q},y_{MT_q}),
\end{equation}
where $P_{MT_q}$ is the probability of predicted by the prompt graph, $y_{MT_q}$ is the corresponding label of query $q$. Overall, the final loss function is as follows:
\begin{equation}
\label{l}
    \mathcal{L} = \mathcal{L}^{NM} + \mathcal{L}^{MT}.
\end{equation}
{Alg~\ref{alg:training-stage} provides the process of the training stage of GraphPrompter, and the Alg~\ref{alg:infer-stage} shows the inference processing.}

\begin{algorithm}[t]
\caption{GraphPrompter: Inference Stage}
\label{alg:infer-stage}
\textbf{Input:} All pre-trained parameters, test Graph $\mathcal{G}$, cache $C$, $k$ (number of selected prompts), $c$ (cache size), the number of test samples $n$ \\
\textbf{Output: } Predicted labels for queries

\begin{algorithmic}[1]
\STATE Initialize all parameters;
\STATE $\mathcal{Y}_{\text{pre}} = \emptyset $
\FOR{step $= 1, 2, \dots, n $}
    \STATE Obtain subgraphs using Prompt Generator, including candidate prompts $\mathcal{S}$ and queries $\mathcal{Q}$;
    \STATE Get their embeddings via pre-trained GNN in Eq~\ref{gi};
    \STATE Compute prompt importance scores using pre-trained selection layers via Eq~\ref{ip};
    \STATE Calculate cosine similarity scores between candidates prompts and queries via Eq~\ref{simpq};
    \STATE Select the top-$k$ prompts based on the combined similarity and importance scores;
    \IF{cache is not empty}
        \STATE Augment the selected prompts by retrieving additional prompts from cache $\mathcal{C}$;
    \ENDIF
    \STATE Update cache $\mathcal{C}$ using the LFU strategy to maintain relevant prompts;
    \STATE Predict the label $y$
    \STATE $\mathcal{Y}_{\text{pre}} = y \cup \mathcal{Y}_{\text{pre}}$
\ENDFOR
\RETURN Predicted labels for queries $\mathcal{Y}_{\text{pre}}$;
\end{algorithmic}
\end{algorithm}
\section{Experiments}

Within this section, we perform comprehensive experiments that encompass node classification and link classification tasks as both pre-training and downstream evaluations. These assessments are carried out on 4 large-scale benchmark datasets with up to 240 million nodes to evaluate the effectiveness of the proposed methods.
\subsection{Experimental Setup}
\subsubsection{\textbf{Datasets}} 
In our study, we adopt two datasets for pre-training, following the approach outlined in Prodigy \cite{huang2023Prodigy}. {The first dataset is MAG240M \cite{hu2021mag240}, which consists of a vast citation network, including over 200 million vertices and more than 100 million edges, making it the largest dataset used in related works \cite{E2GCL, GraphRARE, zhou2024fast}. For such large-scale graph data, we use a batch sampling method to randomly select target nodes and then construct subgraphs using random walks based on the adjacency matrix.} The second dataset is Wiki \cite{xiong-etal-2018-one}, a knowledge graph derived from Wikipedia. Then, we assess the model's in-context learning capabilities using four benchmark graph datasets: arXiv \cite{hu2020arxiv}, ConcepNet \cite{speer2017conceptnet}, FB15K-237 \cite{xiong-etal-2018-one}, and NELL \cite{xiong-etal-2018-one}. {Detailed statistics are presented in Table~\ref{tab:datasets} where the last column refers to the number of node or edge classes.} We utilize subsets from the knowledge graph datasets, as previously employed in \cite{xiong-etal-2018-one}. 

In the case of arXiv, the downstream task entails an $m$-way node classification, predicting the category of the given paper. Conversely, for knowledge graph datasets such as ConceptNet, FB15K-237, and NELL, the downstream task involves an $m$-way relation type classification, projecting the relationship between the two input nodes.

\begin{table}
  \centering
  \caption{Statistics of datasets.}
  \label{tab:datasets}
  \scalebox{1.2}{
  \begin{tabular}{c|c|c|c}
    \toprule
    Dataset & Nodes & Edges & Classes \\
    \midrule
    MAG240M & 244,160,499 & 1,728,364,232 & 153  \\
    Wiki & 4,838,244 & 5,859,240 & 639 \\
    arXiv & 169,343 & 1,166,243 & 40 \\
    ConcepNet & 791k & 2.5M & 14 \\
    FB15K-237 & 14,541 & 310,116 & 200 \\
    NELL & 68,545 & 181,109 & 291 \\
    \bottomrule
  \end{tabular}
  }
\end{table}
\begin{table*}
    \centering
    \caption{\textcolor{black}{The in-context learning accuracy (\%) for arXiv paper category classification is evaluated on 500 sampled test data using 3-shot prompts. Bold indicates the best approach. GraphPrompter was pre-trained on MAG240M and is then applied in-context to arXiv, which has a completely different structure and a different set of paper categories.}}
    \label{tab:arxiv-res}
    \scalebox{1.1}{
        \begin{tabular}{c|cccccc||c}
        \toprule
         Classes  &  NoPretrain &  Contrastive &  Finetune & Prodigy & ProG & OFA & GraphPrompter \\
    \midrule
         3&  33.16 \small{$\pm$0.30} & 65.08 \small{$\pm$0.34}  &  65.42 \small{$\pm$5.53} & 73.09 \small{$\pm$0.36} & 50.0 \small{$\pm$0.0} & 72.24 \small{$\pm$3.81} & \textbf{78.57 \small{$\pm$15.21}}  \\
         5& 18.33 \small{$\pm$0.21} & 51.63 \small{$\pm$0.29} & 53.49 \small{$\pm$4.61} & 61.52 \small{$\pm$0.28} & 58.33 \small{$\pm$8.33} & 59.78\small{$\pm$2.51} & \textbf{68.85 \small{$\pm$13.05}} \\
         10& 9.19 \small{$\pm$0.11} & 36.78 \small{$\pm$0.19} & 30.22 \small{$\pm$3.77}  & 46.74 \small{$\pm$0.20}  & 53.33 \small{$\pm$6.67} & - & \textbf{54.53 \small{$\pm$10.12}}  \\
         20& 4.72 \small{$\pm$0.06} & 25.18 \small{$\pm$0.11}  & 17.68 \small{$\pm$1.15} & 34.41 \small{$\pm$0.12} &  35.0 \small{$\pm$22.91} & - & \textbf{40.74 \small{$\pm$6.68}}\\
         40& 2.62 \small{$\pm$0.02} &  17.02 \small{$\pm$0.07} & 8.04 \small{$\pm$3.00} & 25.13 \small{$\pm$0.07} &  \textbf{47.50 \small{$\pm$6.67}} & - & {29.47 \small{$\pm$3.82}}  \\

    \bottomrule
    \end{tabular}}
    
\end{table*}

\begin{table*}
    \centering
    \caption{\textcolor{black}{On 500 sampled test data, the in-context learning accuracy (\%) evaluations are conducted on ConceptNet, FB15K-237, and NELL datasets (ordered from top to bottom) using 3-shot prompts. Bold indicates the best approach. GraphPrompter was pre-trained on Wiki, which has completely different node and relation types from graphs it is then applied on in-context.}}
    \label{tab:edge-res}
    \scalebox{1.1}{
    \begin{tabular}{c|cccccc|| c}
        \toprule
         Classes  &  NoPretrain &  Contrastive &  Finetune & Prodigy & ProG & OFA & GraphPrompter \\
    \midrule
         4 &  30.40 \small{$\pm$0.63} & 44.01 \small{$\pm$0.61}  &  53.85 \small{$\pm$9.29} & 53.97 \small{$\pm$0.63} & - & - & \textbf{58.46 \small{$\pm$15.2}}  \\
         \midrule
         5 & 33.54 \small{$\pm$0.61} & 81.35 \small{$\pm$0.58} & 82.01 \small{$\pm$12.83} & 88.02 \small{$\pm$0.48}  & 41.67 \small{$\pm$8.33} & - &\textbf{99.65 \small{$\pm$1.31}} \\
         10& 20.00 \small{$\pm$0.35} & 70.88 \small{$\pm$0.48} & 71.97 \small{$\pm$6.16}  & 81.10 \small{$\pm$0.39}   & 43.33 \small{$\pm$13.33} & \textbf{89.68 \small{$\pm$1.38}} & {89.52 \small{$\pm$4.07}} \\
         20& 9.20 \small{$\pm$0.18} & 59.80 \small{$\pm$0.35}  & 64.01 \small{$\pm$4.66} & 72.04 \small{$\pm$0.33}  & 50.00 \small{$\pm$12.90} & 81.33 \small{$\pm$2.54} &\textbf{83.78 \small{$\pm$3.51}}  \\
         40 & 2.5 \small{$\pm$0.08} &  49.39 \small{$\pm$0.23} & 57.27 \small{$\pm$3.33} & 59.58 \small{$\pm$0.22}  & 34.17 \small{$\pm$20.05} & - & \textbf{66.94 \small{$\pm$3.12}} \\
         \midrule
         5 & 33.44 \small{$\pm$0.57} & 84.08 \small{$\pm$0.54} & 87.22 \small{$\pm$12.75} & 87.02 \small{$\pm$0.44}  & 75.00 \small{$\pm$25.00} & - & \textbf{93.34 \small{$\pm$5.37}} \\
         10 & 18.82 \small{$\pm$0.31} & 76.54 \small{$\pm$0.45} & 71.90 \small{$\pm$5.90}  & 81.06 \small{$\pm$0.41}   & 44.00 \small{$\pm$12.00} & - & \textbf{87.47 \small{$\pm$4.83}}  \\
         20 & 7.42 \small{$\pm$0.16} & 66.56 \small{$\pm$0.35}  & 66.19 \small{$\pm$8.46} & 72.66 \small{$\pm$0.32}  & 47.67 \small{$\pm$31.69} & - & \textbf{81.46 \small{$\pm$4.09}}  \\
         40 & 3.04 \small{$\pm$0.07} &  57.44 \small{$\pm$0.24} & 55.06 \small{$\pm$4.19} & 60.02 \small{$\pm$0.22}  & 47.16 \small{$\pm$22.37} & - & {\textbf{75.74 \small{$\pm$2.86}}} \\
    \bottomrule
    \end{tabular}}
\end{table*}

\subsubsection{\textbf{Evaluation}} We evaluate the in-context learning performance on various downstream datasets within the same domain as the pre-training dataset. For instance, we pre-train on Wiki and evaluate on ConceptNet, FB15K-237, and NELL. Each downstream classification dataset is accompanied by its original train, validation, and test partitions. 

To simulate the scenarios where labeled data is limited in the downstream task, we select $N$ ($=10$) nodes or edges from the training partition per category as candidate prompt examples with known labels. Subsequently, we construct a $k$-shot $l$  ($=1$) prompt for test nodes or edges from the test partition, by selecting $k$ ($=3$) examples per category from these available instances. 
Unlike Prodigy, which chooses both candidate prompt examples and prompts randomly, we employ our proposed multi-stage adaptive approach for their selection. All our pre-training and testing settings are consistent with Prodigy. Therefore, we extract some results from Prodigy for comparison.

\begin{table}[t]
  \caption{{The in-context learning accuracy (\%) on FB15K-237 and NELL datasets with 3-shots, which consit of 50,60,80 and 100 classes respectively. Bold indicates the best approach.}}
  \label{tab:more-ways}
  \scalebox{0.83}{
  \begin{tabular}{c|ccccc}
    \toprule
     & \multicolumn{4}{c}{FB15K-237} \\
    Classes  & 50 & 60 & 80 & 100 \\
    \midrule
    Prodigy  & 55.34 \small{$\pm$4.01} & 49.54 \small{$\pm$3.81} & 37.06 \small{$\pm$2.95} & 27.39\small{$\pm$2.34} \\
    ProG & 34.66 \small{$\pm$21.40} & 48.33 \small{$\pm$21.23}& 30.65 \small{$\pm$24.9} & 25.23 \small{$\pm$18.42} \\
    \midrule
    GraphPrompter & \textbf{62.735 \small{$\pm$2.98}} & \textbf{53.95 \small{$\pm$2.83}}  & \textbf{42.96 \small{$\pm$2.62}} & \textbf{28.03\small{$\pm$2.17}} \\
    \midrule
    & \multicolumn{4}{c}{NELL} \\
    Classes  & 50 & 60 & 80 & 100 \\
    \midrule
    Prodigy  & 56.72 \small{$\pm$4.08} & 50.25 \small{$\pm$3.86}  & 40.64 \small{$\pm$4.94} & 28.47 \small{$\pm$3.64}  \\
    ProG & 22.0 \small{$\pm$20.39} & 22.78 \small{$\pm$24.14} & 26.54 \small{$\pm$19.26} & 24.13 \small{$\pm$20.05} \\
    \midrule
    GraphPrompter  & \textbf{66.36 \small{$\pm$2.91}} & \textbf{61.16 \small{$\pm$2.78}} & \textbf{53.73 \small{$\pm$5.13}} & \textbf{35.95 \small{$\pm$4.42}}  \\
    \bottomrule
  \end{tabular}}
\end{table}

\begin{table}[t]
\caption{{The in-context learning accuracy (\%) on arXiv and FB15K-237 datasets with 3-shots.}}
  \label{tab:ofa-c}
  \scalebox{0.8}{
  \begin{tabular}{c|ccccc}
    \toprule
     & \multicolumn{4}{c}{arXiv}  \\
    Classes  & 3 & 5 & 10 & 20 \\
    \midrule
    OFA  & 46.16 \small{$\pm$14.84} & 32.73 \small{$\pm$15.15} & 19.8 \small{$\pm$11.69} & 12.03 \small{$\pm$8.32} \\
    GraphPrompter & \textbf{78.57 \small{$\pm$15.21}} & \textbf{68.85 \small{$\pm$13.05}}& \textbf{54.53 \small{$\pm$10.12}} & \textbf{40.74 \small{$\pm$6.68}} \\
    \midrule
     & \multicolumn{4}{c}{FB15K-237} \\
    Classes  & 5 & 10 & 20 & 40 \\
    \midrule
    OFA  & 75.43 \small{$\pm$27.38} & 65.67 \small{$\pm$33.11} & 55.56 \small{$\pm$35.42} & 45.17 \small{$\pm$33.17} \\
    GraphPrompter & \textbf{99.65 \small{$\pm$1.31}} & \textbf{89.52 \small{$\pm$4.07}}& \textbf{83.78 \small{$\pm$3.51}} & \textbf{66.94 \small{$\pm$3.12}} \\
    \bottomrule
  \end{tabular}}

\end{table}

\subsubsection{\textbf{Baselines}}
We consider four baseline models for comparative analysis: 
\begin{itemize}
    \item NoPretrain: This baseline employs a model with the same architecture as the pre-trained models, but with randomly initialized weights.
    \item{Contrastive}~\cite{you2020graph}: Utilizing a standard contrastive learning approach with self-supervised augmentation, this baseline adapts to the in-context learning setting by implementing a hard-coded nearest neighbor algorithm. More specifically, we classify the query by comparing its pre-trained embedding against the average embedding of the example inputs for each class. 
    \item{Finetune}~\cite{Hu*2020Strategies}: This baseline extends upon the graph encoder pre-trained with contrastive learning by training an additional linear classification head, following common practice. 
    \item{Prodigy}~\cite{huang2023Prodigy}: This baseline randomly selects both candidate prompt examples and prompts, without employing any augmentation strategy during the testing stage. 
    \item{All-in-One (ProG)}~\cite{allinone}: The baseline introduces an additional learnable subgraph as a prompt, which uses meta-learning to tune the prompt in testing.
    \item{One-For-All (OFA)}~\cite{oneforall}: This baseline describes nodes and edges with natural language uses LLMs to encode the diverse and possibly cross-domain text attributes, and uses text feature nodes to construct prompt graph. Following our few shots and cross-domain tasks, we use the OFA low resource version (OFA-joint-lr), which is in the cross-domain settings. As the OFA's prediction results are unstable due to dataset partitioning in the few-shot scenario\footnote{\url{https://github.com/LechengKong/OneForAll/issues/11}}, we extract the results from OFA claimed for comparison, and we present our test results in~\ref{tab:ofa-c}.
\end{itemize}
\subsubsection{\textbf{Models configurations}} We follow the Prodigy\cite{huang2023Prodigy} settings to pre-trained the graph models.
\begin{itemize}
    \item {Training, MAG240M:}
    The pretraining task we used consisted of 30 ways, 3 shots, and 4 queries per task. This specific task configuration was carefully selected to strike a balance between complexity and diversity in the training data, without overwhelming the GPU memory. We checkpoint the model every 500 steps. 
    \item {Training, Wiki:}
    Our pretraining setup included a model with an input dimension of 768 and an embedding dimension of 256, the AdamW optimizer with a learning rate of $1\times10^{-3}$ and weight decay of $1\times10^{-3}$, the pretraining task settings are similar with MAG240M.    
    \item {GNN architecture:}
    We use GraphSAGE\cite{hamilton2017inductive} to generate the embeddings for data graph prompts in Eq~\ref{gi}, which has been proven to have good scalability on large-scale graphs. For the Task Graph, we use the attention-based graph model following Prodigy~\cite{huang2023Prodigy}.
\end{itemize}


\subsection{Main Results}
Following the Prodigy approach, we commence our evaluation by assessing the in-context learning efficacy in node classification and link prediction tasks, considering different numbers of ways (referring to the number of classes to be classified among). In the edge classification task, we employed the online prompts augmentation strategy to enhance the generalization capability of the pre-trained model on the test dataset.

\subsubsection{\textbf{Enhanced In-Context Learning Performance}}
As illustrated in Table~\ref{tab:arxiv-res} and Table~\ref{tab:edge-res}, the best score is \textbf{Bolded}, our approach consistently surpasses all other baselines in this particular scenario, showcasing stronger in-context learning performance. On the arXiv dataset, it achieves the highest average accuracy across most ways, improving by 48\% on average and up to 54.43\% over the best-performing Prodigy baseline. On KG datasets, GraphPrompter raises Prodigy’s score from 68.4\% to 81.82\%. Furthermore, our method excels in comparison to Finetune, which necessitates additional training on downstream tasks. On the arXiv dataset, we observe an average improvement of 19.6\% across all ways. Similarly, on KG datasets, we achieve over 15\% improvements across all ways, surpassing the performance of all classes, even in settings where Prodigy does not perform well (e.g., NELL-20-way). This can be attributed to Prodigy's incorporation of a diverse range of pretraining tasks, enabling the model to avoid overfitting on specific tasks and acquire contextual knowledge. We also compare with Contrastive \cite{you2020graph}, which adapts a standard contrastive learning method using a hard-coded nearest-neighbor classifier. Our method outperforms it by over 14\% on KGs, indicating that while simple nearest-neighbor approaches are not optimal for in-context learning, they still retain some effectiveness.

In fact, we can prove that, hard-code nearest neighbor may not perform well to learning in-context information, but it is not just useless, it can be adaptively integrated into Graph models and enhance the graph in-context learning ability.  In the same few-shot scenario, the performance of the ProG~\cite{allinone} in cross-domain tasks is average 48.24\%. This can be understood as the learnable prompt-based methods requiring more samples for fine-tuning the prompt tokens. {Although ProG shows a brief improvement on 10 to 20 classes, but performance drops as the number of classes increases (see Table~\ref{tab:more-ways}). We attribute this to ProG's fine-tuning, which slightly improves with larger data volumes. However, its performance still lags behind ours and demonstrates instability in predictions.} For OFA~\cite{oneforall}, which has a similar prompt structure to Prodigy, we also achieved better performance on the arXiv and FB15K-237 datasets. This suggests that the prompt optimization strategy we proposed has great application potential in methods related to prompt graphs. 
Table \ref{tab:ofa-c} presents the comparison of experimental results between our GraphPrompter and OFA \cite{oneforall} under the same settings. For OFA, we use the same random category selection method as our experiment and employ the OFA-joint-lr approach (which trains and evaluates a single model using all datasets simultaneously) to align with our cross-domain task. 
The experimental results show that our method performs more stably and better when randomly selecting categories.
Furthermore, we analyze the inference time, shown in the Table~\ref{tab:time}. The experimental results indicate that our method can achieve favorable outcomes without introducing a significant burden.

\begin{figure}[t]

  \centering
  \includegraphics[width=1\linewidth]{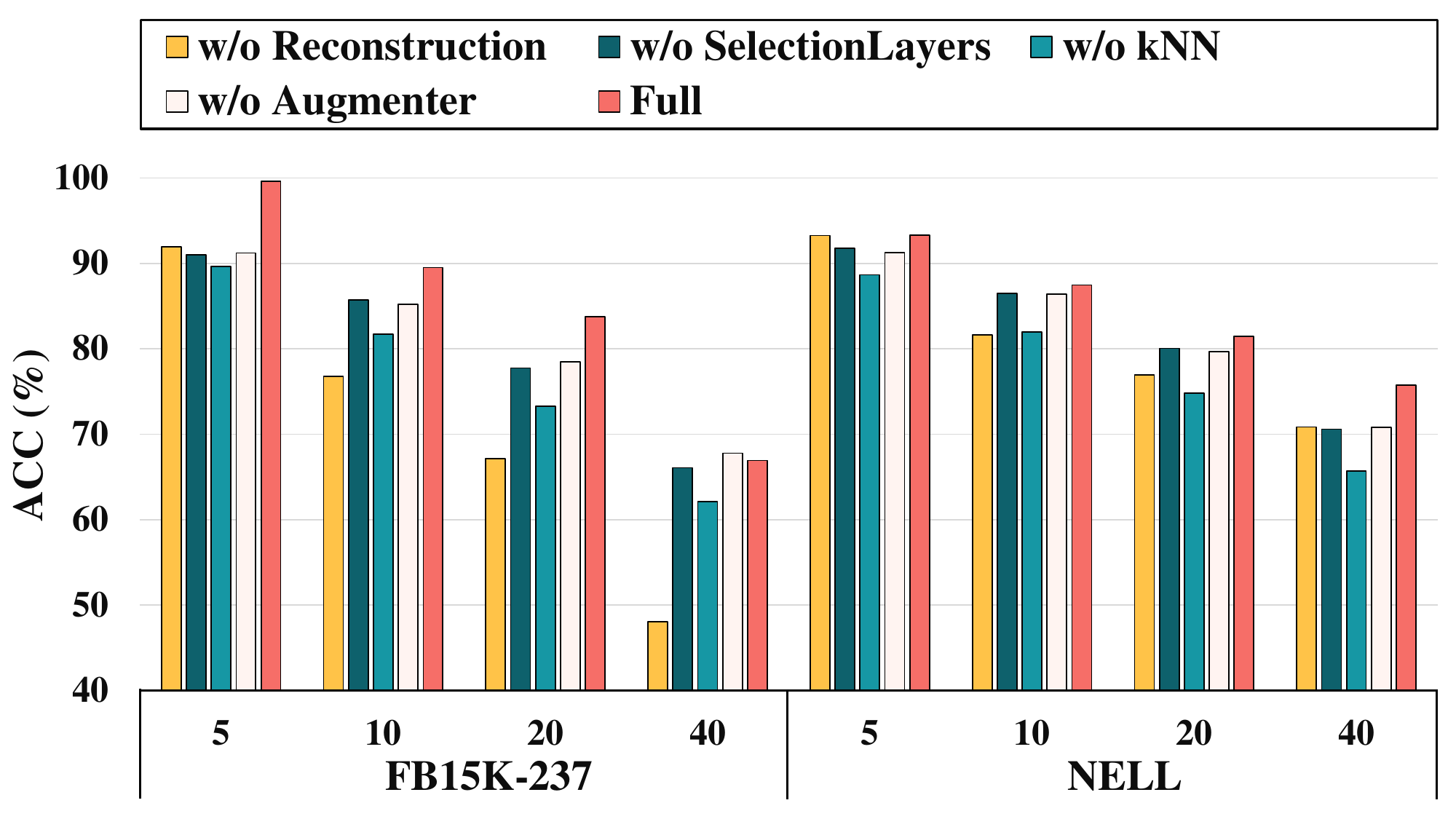}
  \caption{\textcolor{black}{Ablation study on FB15K-237 and NELL with 3-shots wrt. the number of ways from 5 to 40.}}
  \label{fig:ablation-study}
\end{figure}

\begin{figure}

  \centering
  \includegraphics[width=1\linewidth]{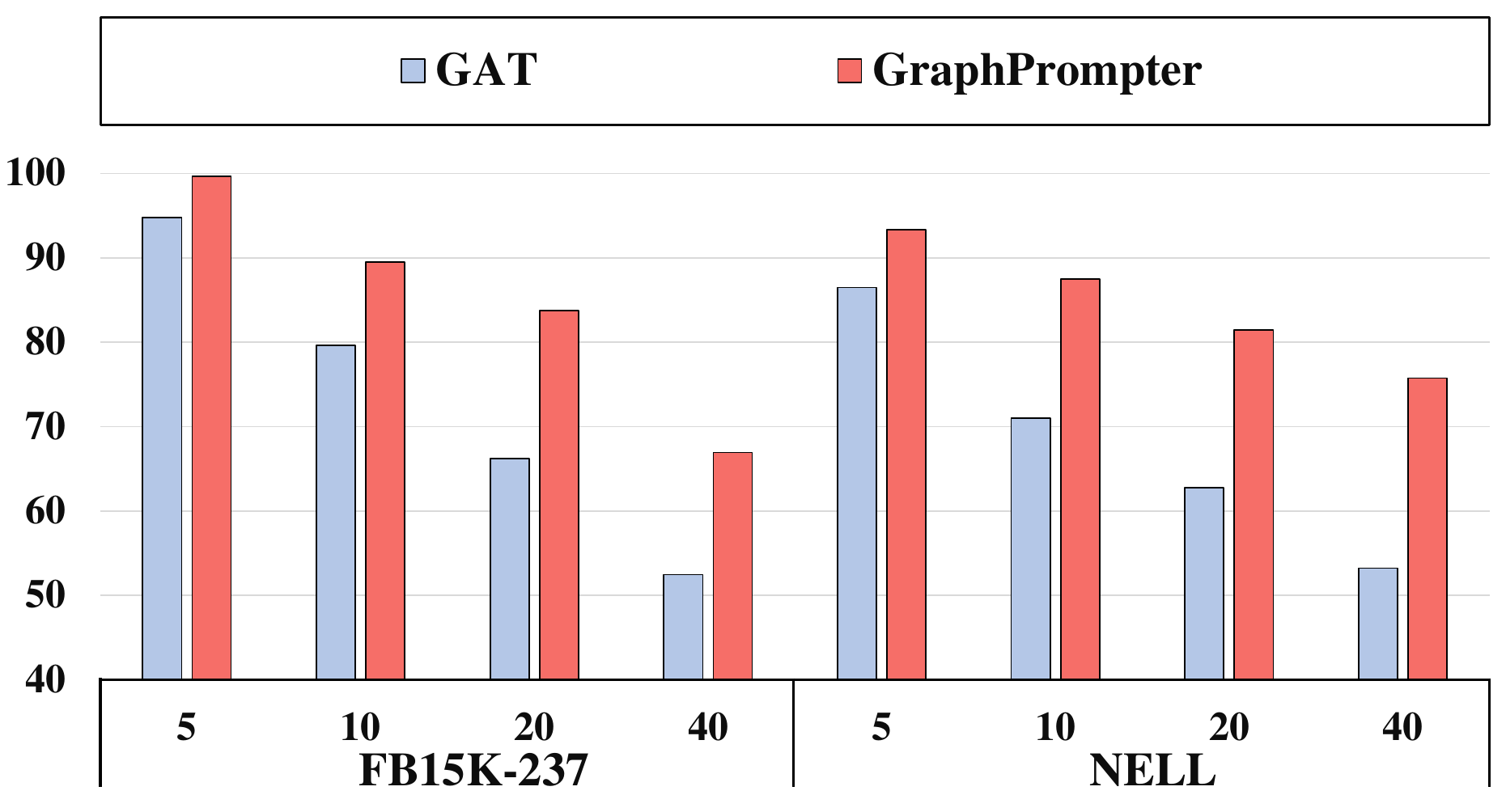}
  \caption{\textcolor{black}{The comparison of different GNN architectures on FB15K-237 and NELL datasets.}}
  \label{fig:gat}
\end{figure}

\begin{figure}[t]
  \centering
  \includegraphics[width=1\linewidth]{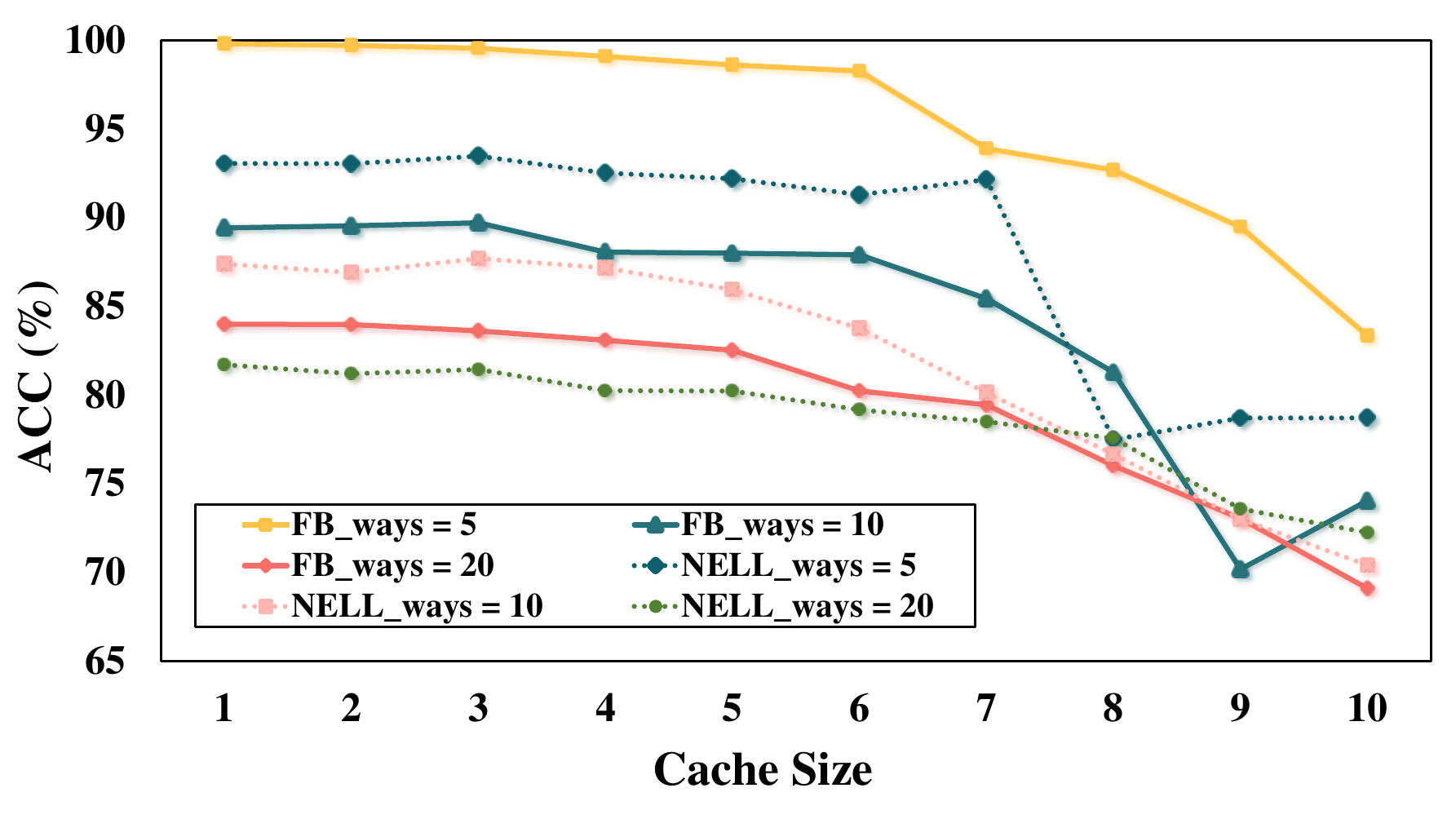}
  \caption{\textcolor{black}{Analysis of cache size from 1 to 10. FB noted FB15K-237 dataset and NELL means NELL dataset. }}
  \label{fig:cache}
\end{figure}

\begin{table}[t]
  \caption{{Random select sample as a pseudo-label using different random seeds on FB15K-237-20ways and NELL-20ways dataset.}}
  \label{tab:cache}
  \scalebox{0.96}{\begin{tabular}{c|ccccccc}
    \toprule
    random-seed & 10 & 30 & 50 & 70 & 90 & Avg. $\pm$ std.\\
    \midrule
    FB15K-237  & 79.98 & 82.05 & 82.01 & 78.93 & 80.34 & 80.66 $\pm$ 1.21 \\
    \midrule
    NELL & 80.95 & 80.47 & 76.68 & 78.67 & 79.89 & 79.33 $\pm$ 1.53 \\
    \bottomrule
  \end{tabular}} 
\end{table}

\begin{figure}[t]
      \centering 
      \scalebox{0.95}{\subfloat{
      \includegraphics[width=0.5\linewidth]{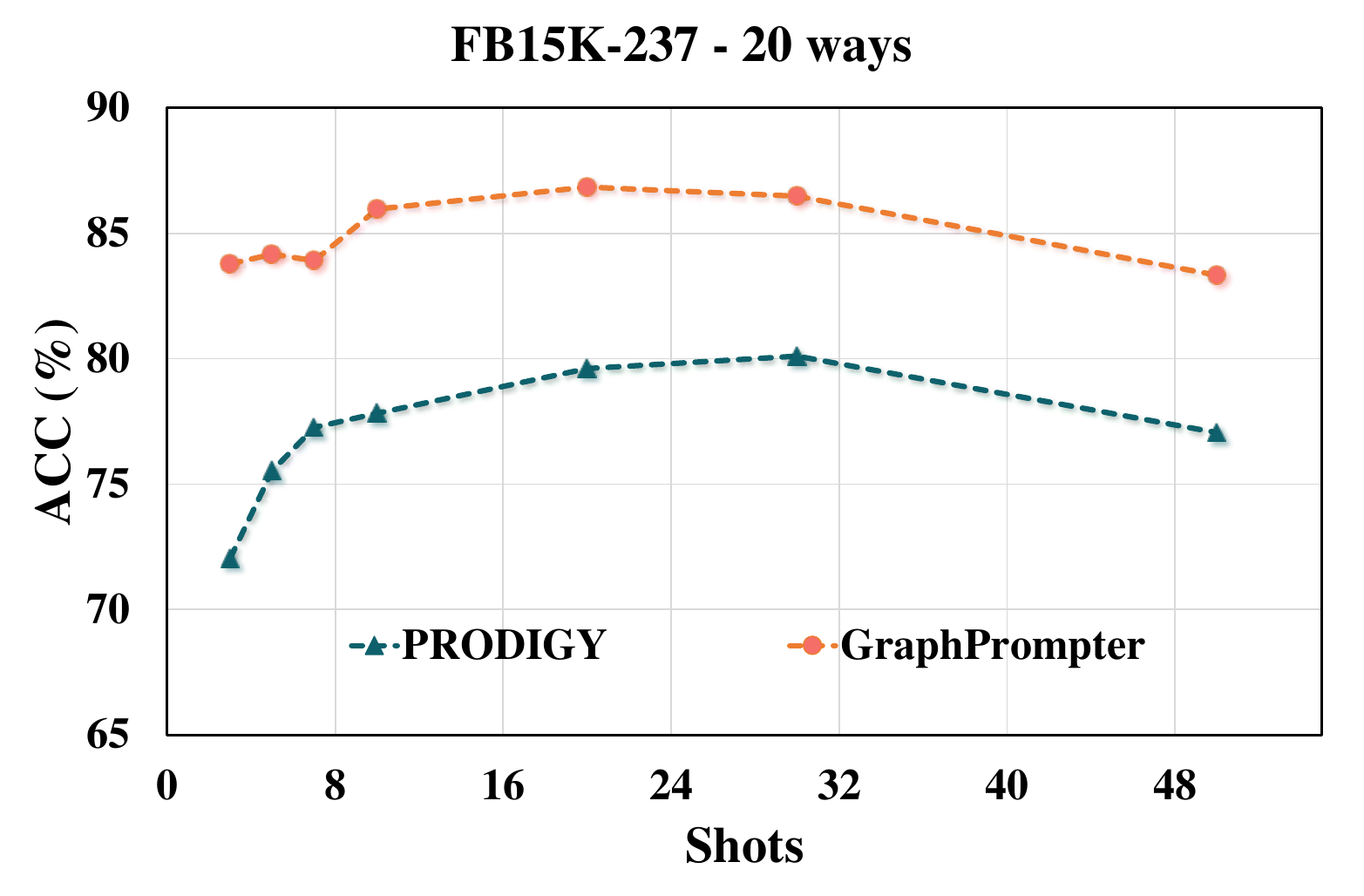}
      }
      \subfloat{
      \includegraphics[width=0.5\linewidth]{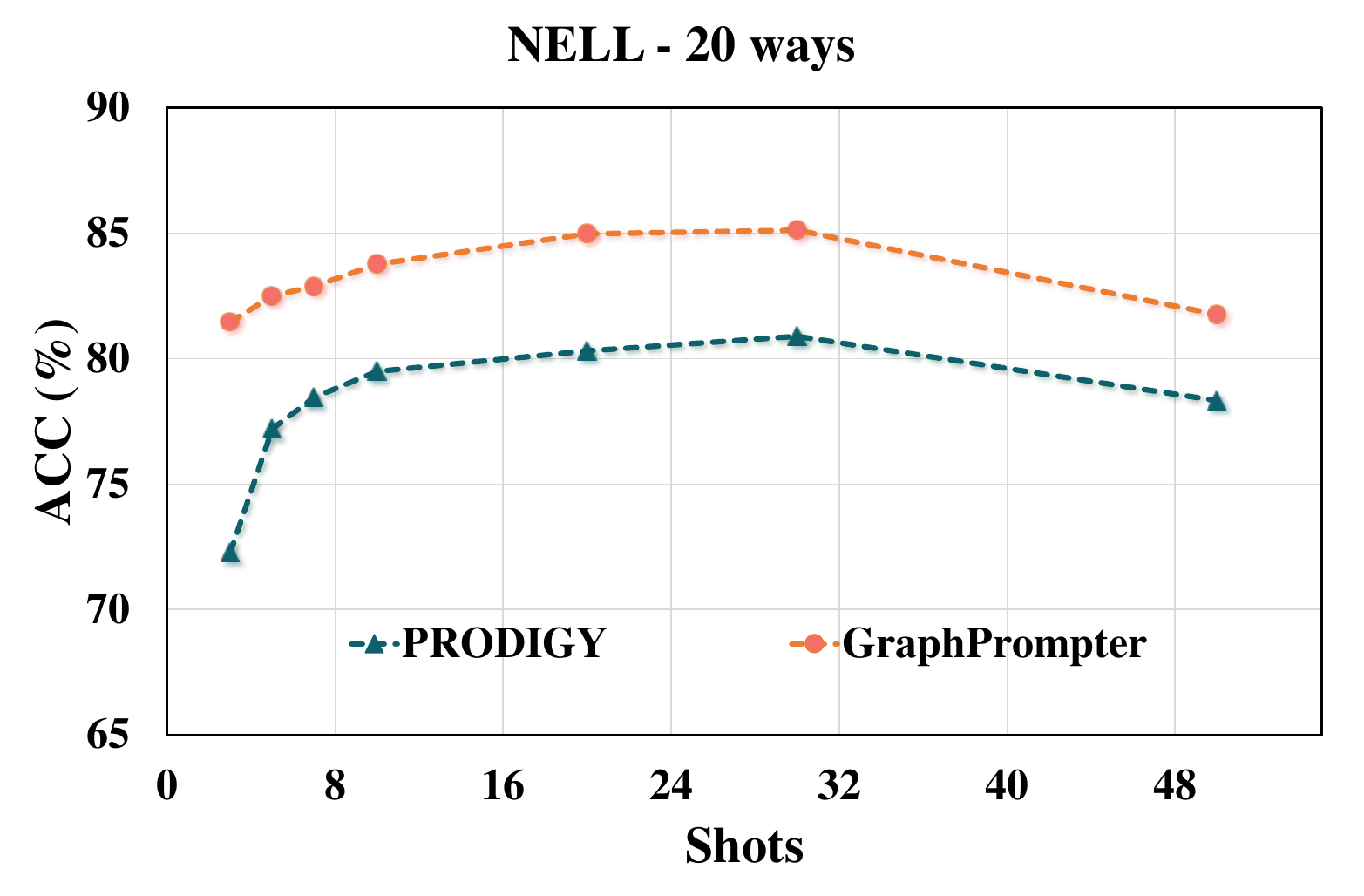}
      }}
      \\
      \scalebox{0.95}{\subfloat{
      \includegraphics[width=0.5\linewidth]{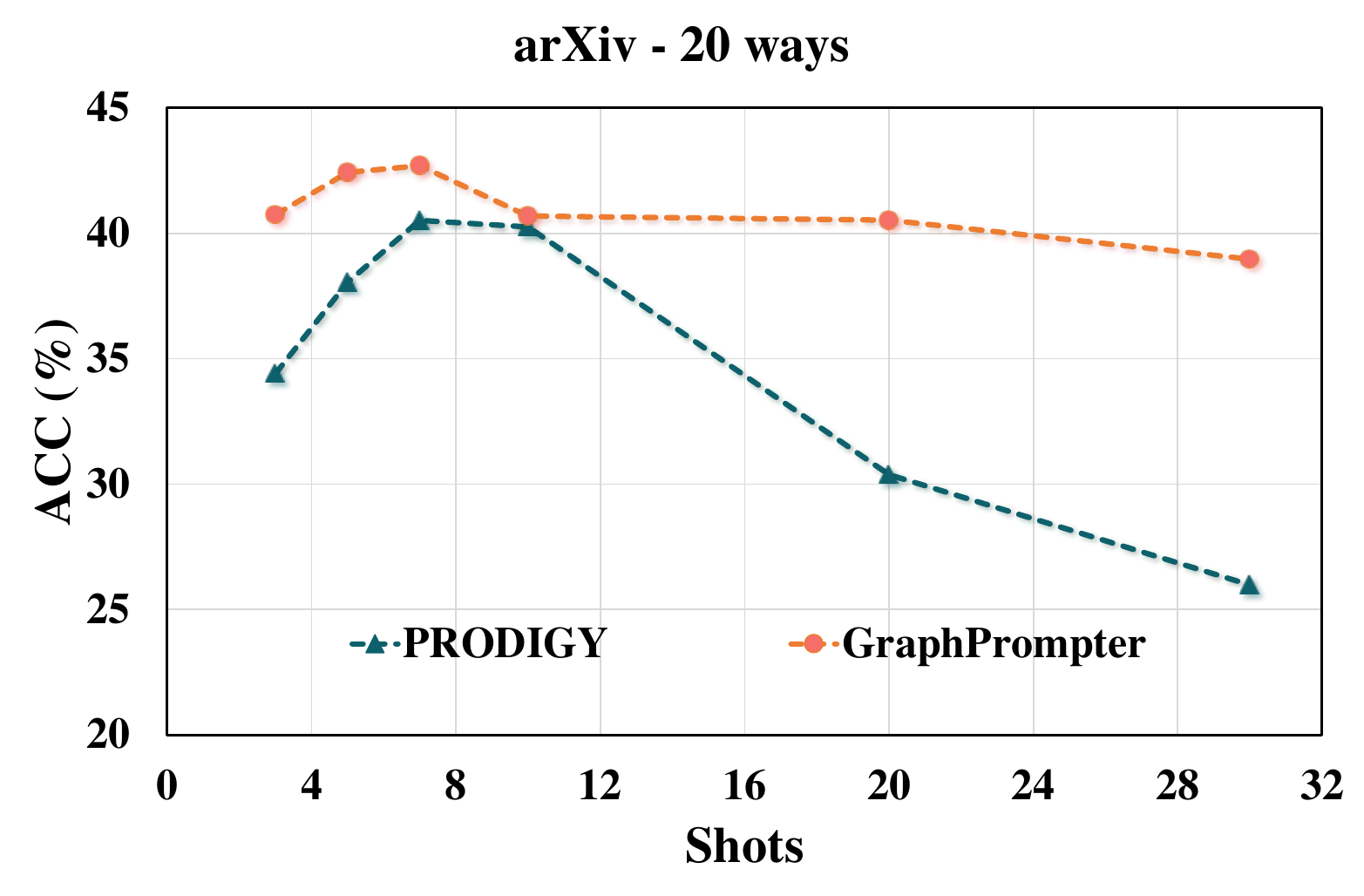}
      }
      \subfloat{
      \includegraphics[width=0.5\linewidth]{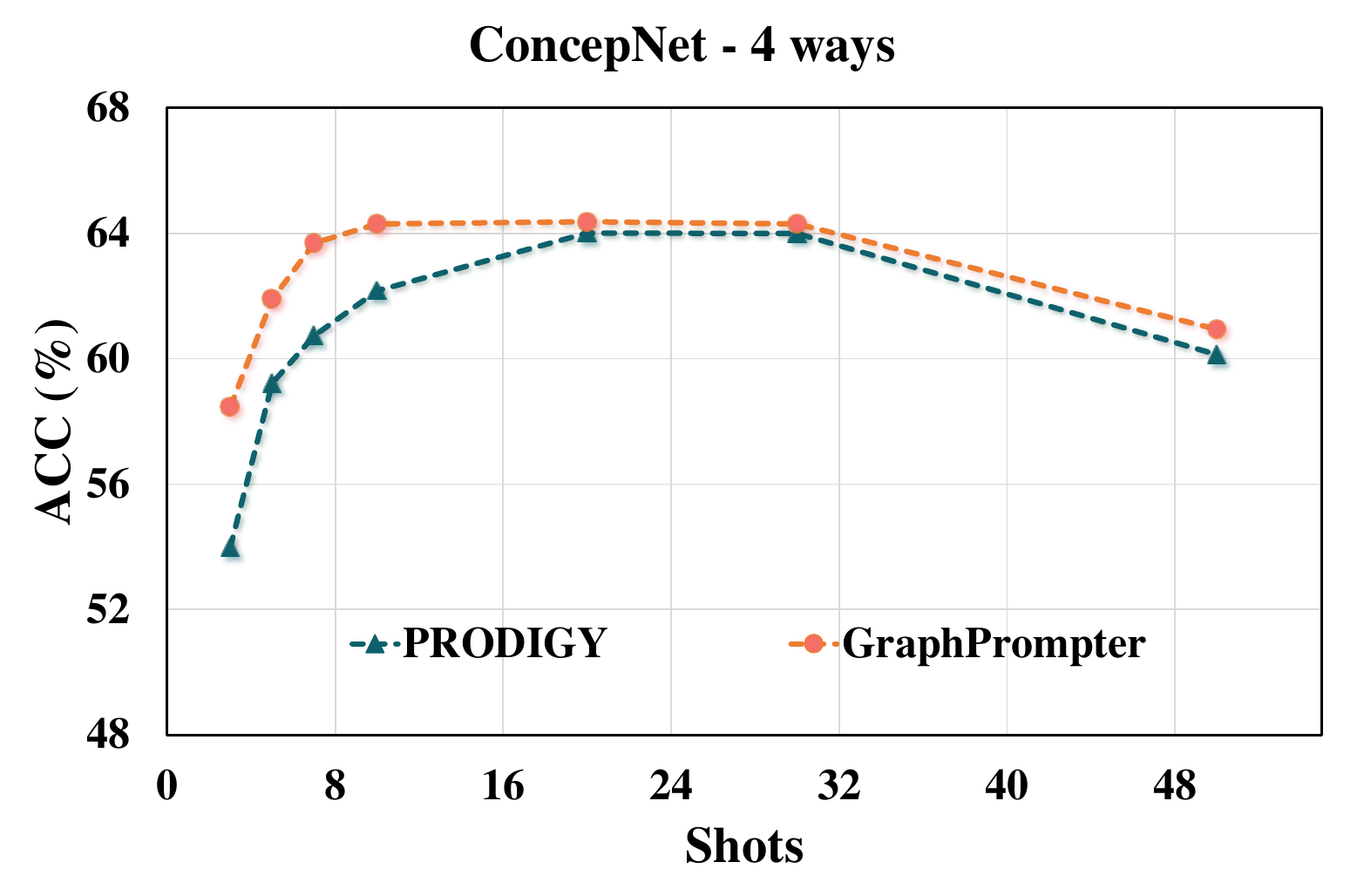}
      }}
  
      \caption{In-context learning accuracy on FB15K-237, NELL, arXiv, and ConcepNet datasets wrt. the number of prompt examples(shots), respectively.}
      \label{fig:diff-prompts}

  \end{figure}

\begin{figure*}[t]
      \centering 
     \scalebox{0.95}{ \subfloat[Prodigy - 20 shots on NELL]{
      \includegraphics[width=0.25\linewidth]{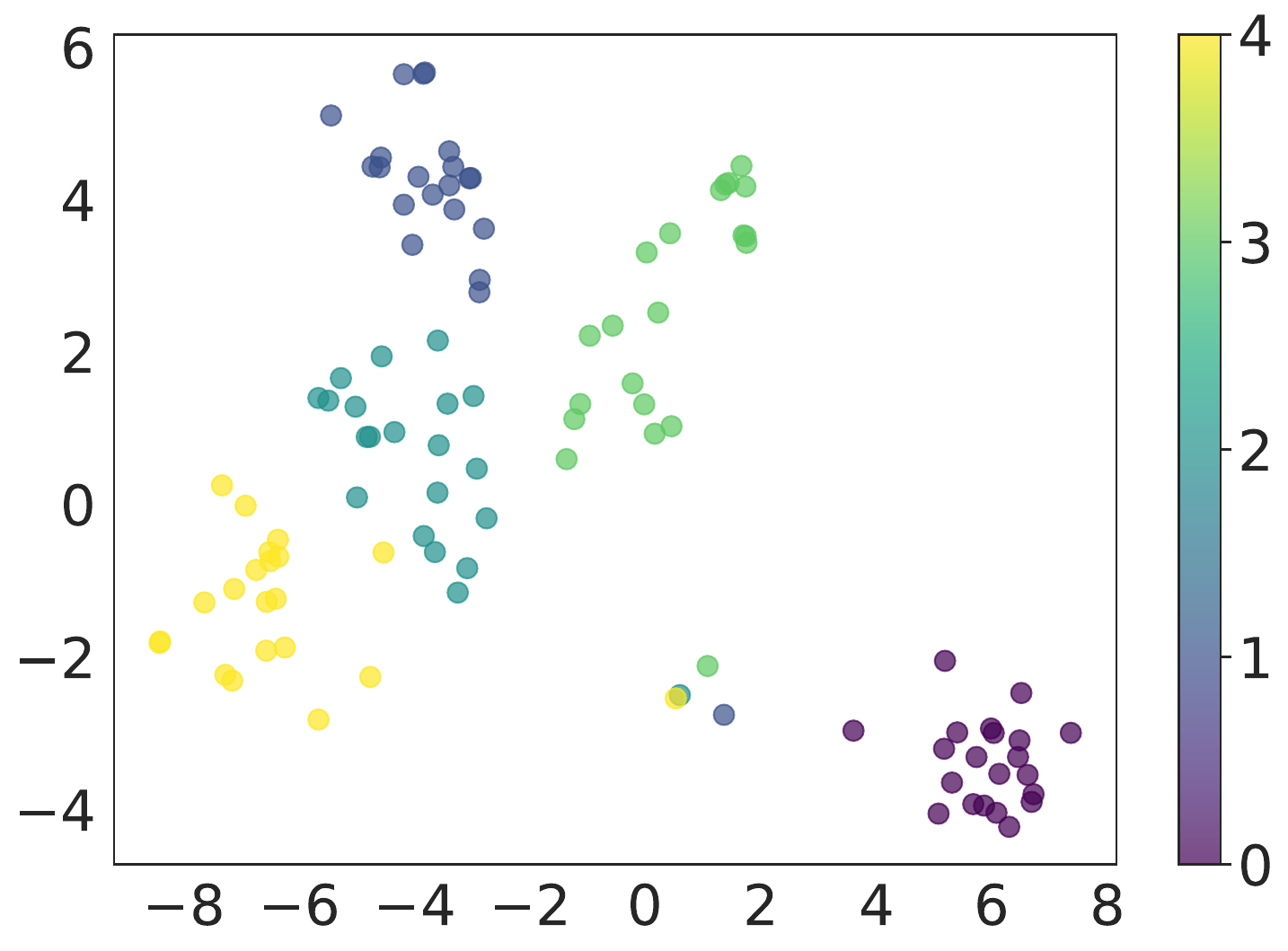}
      }
      \subfloat[GraphPrompter - 20 shots on NELL]{
      \includegraphics[width=0.25\linewidth]{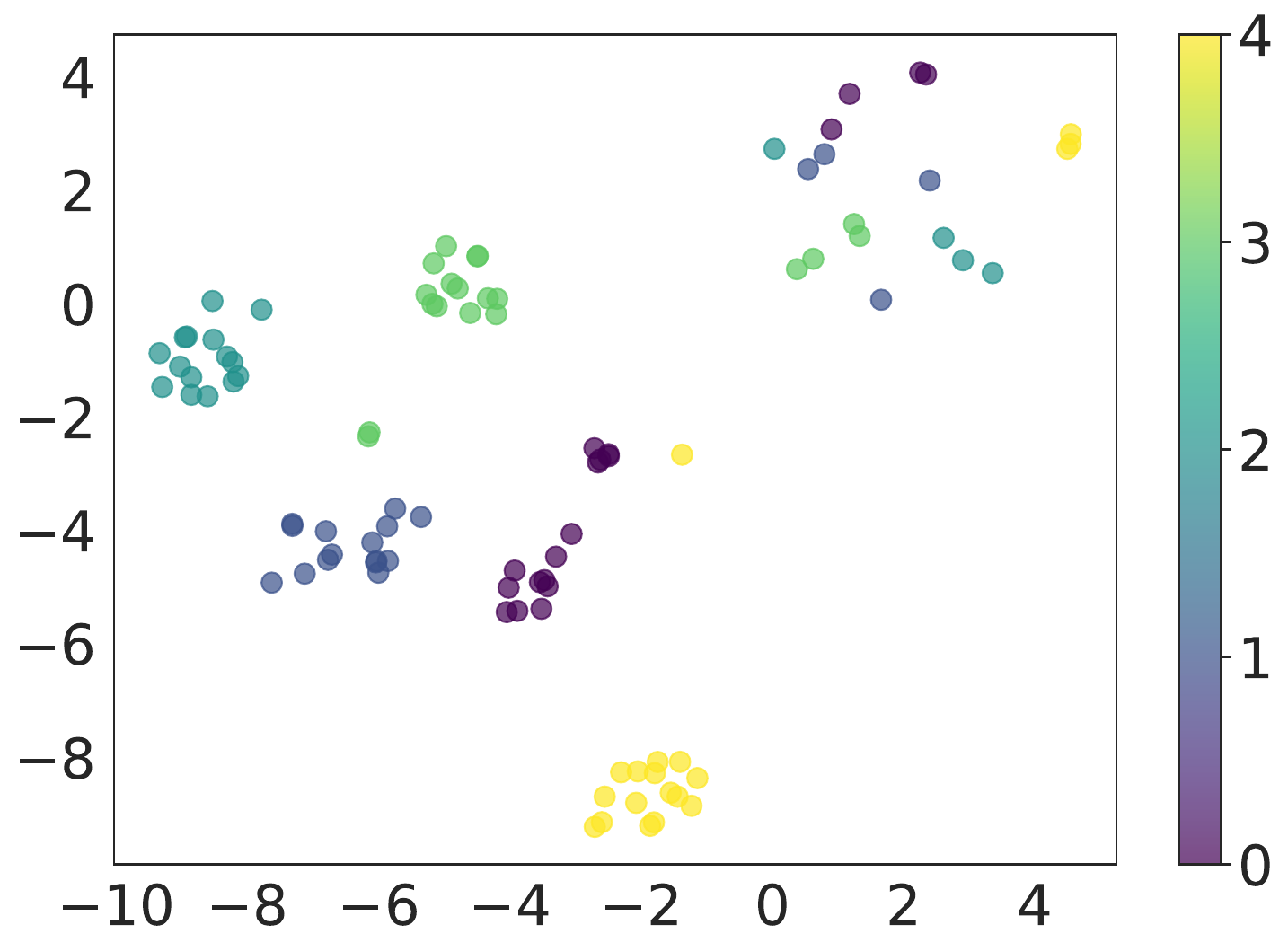}
      }
      \subfloat[Prodigy - 50 shots on NELL]{
      \includegraphics[width=0.25\linewidth]{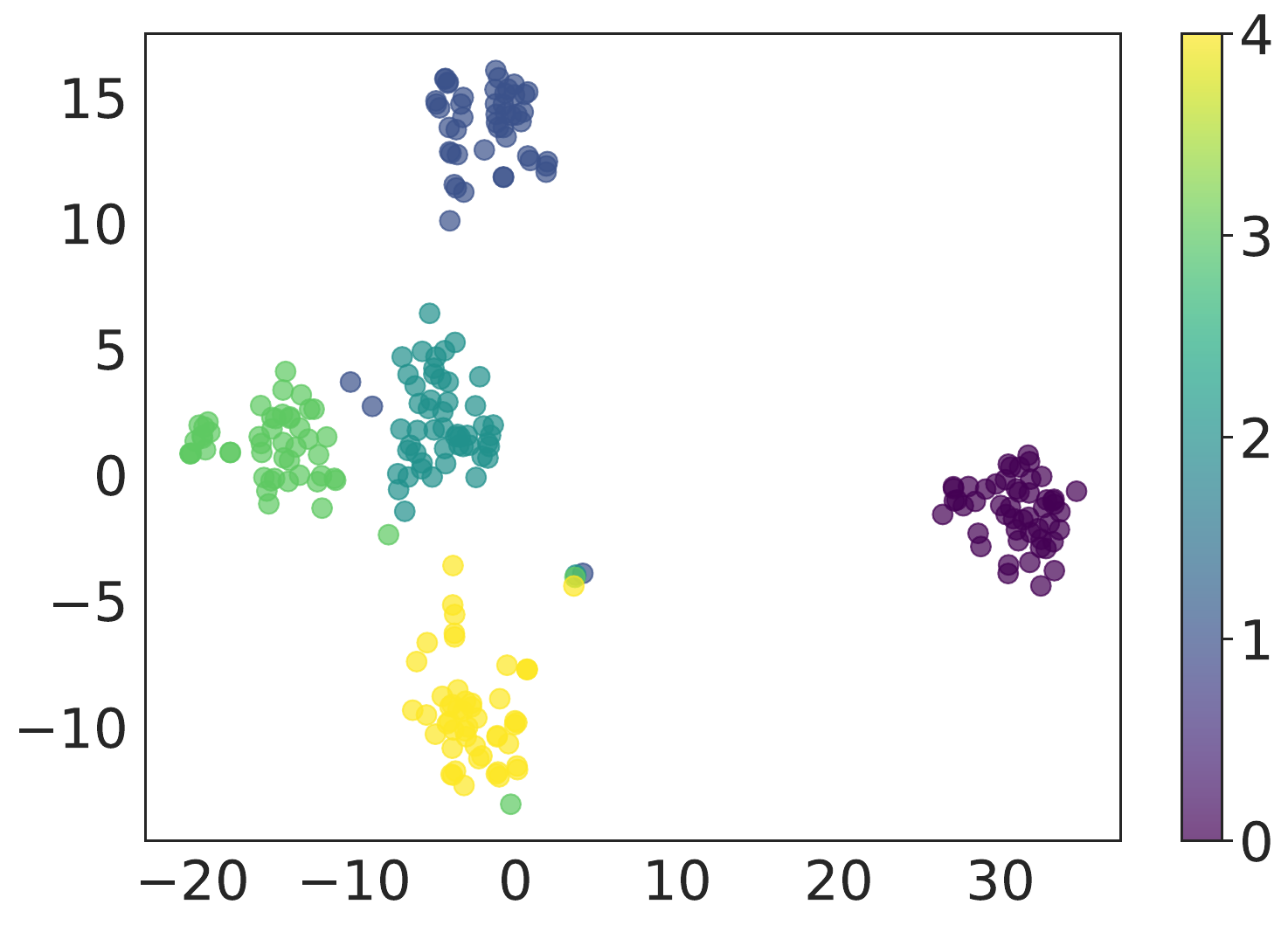}
      }
      \subfloat[GraphPrompter - 50 shots on NELL]{
      \includegraphics[width=0.25\linewidth]{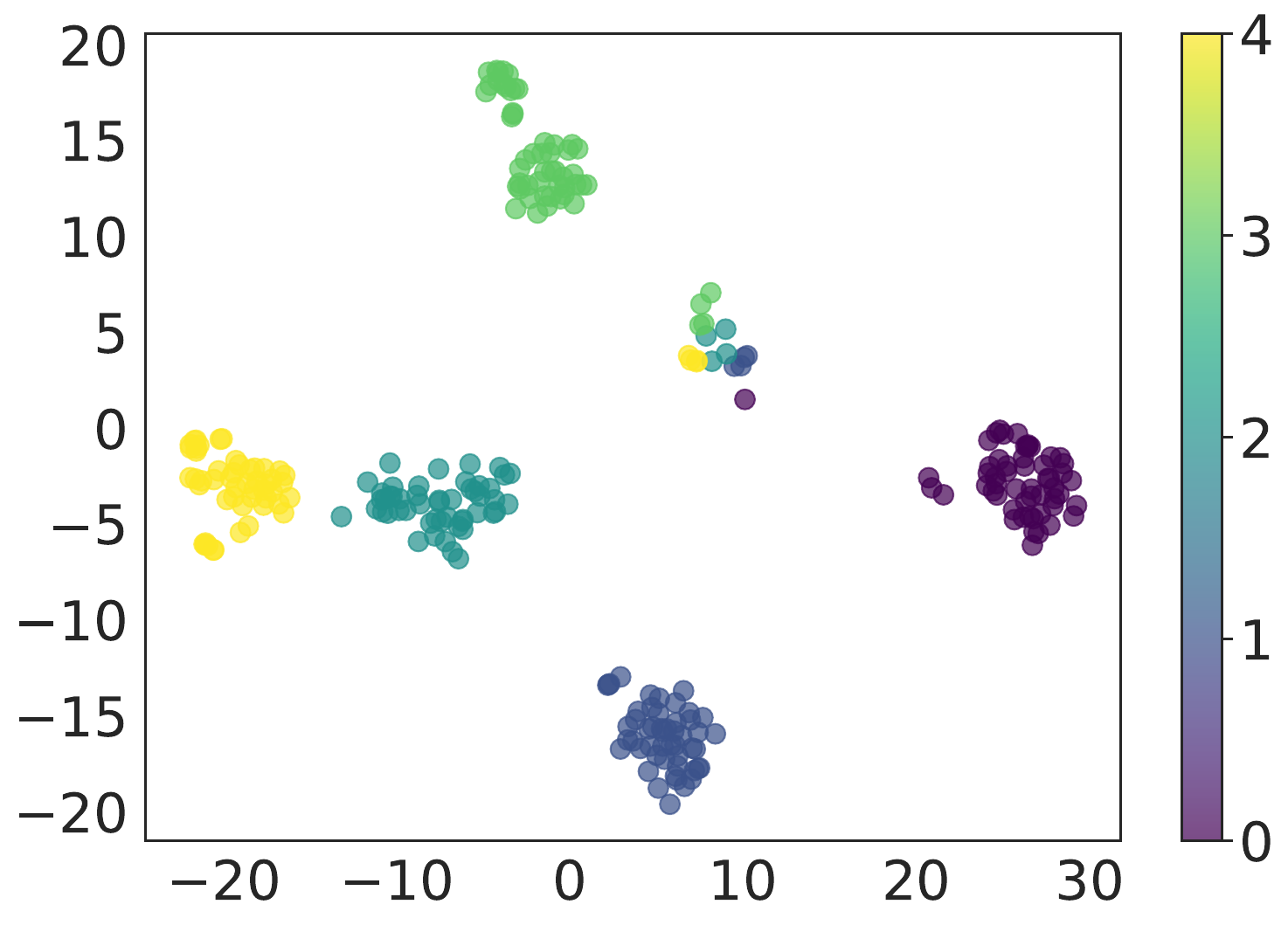}
      }}
      \\
      \scalebox{0.95}{\subfloat[Prodigy - 20 shots on FB]{
      \includegraphics[width=0.25\linewidth]{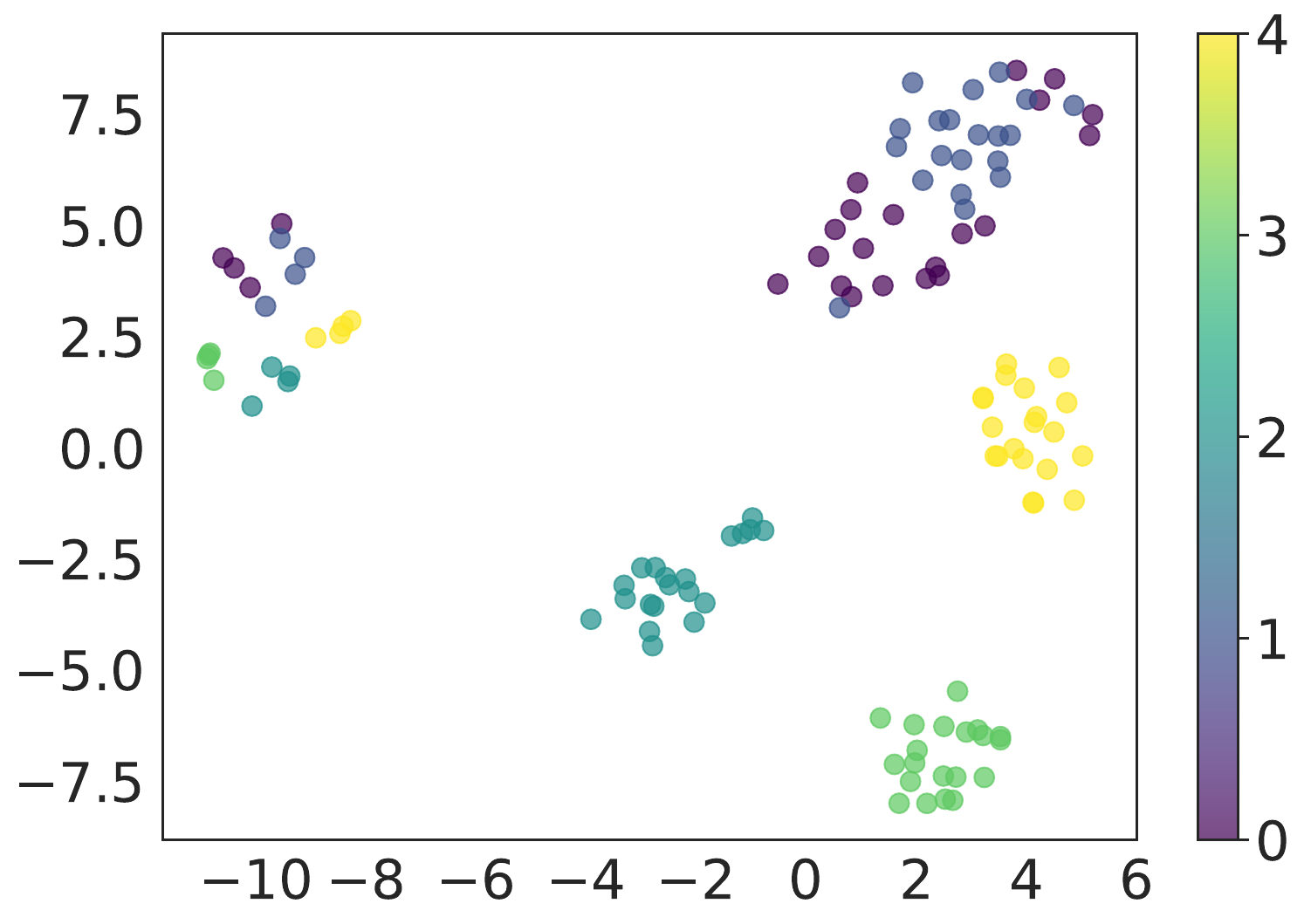}
      }
      \subfloat[GraphPrompter - 20 shots on FB]{
      \includegraphics[width=0.25\linewidth]{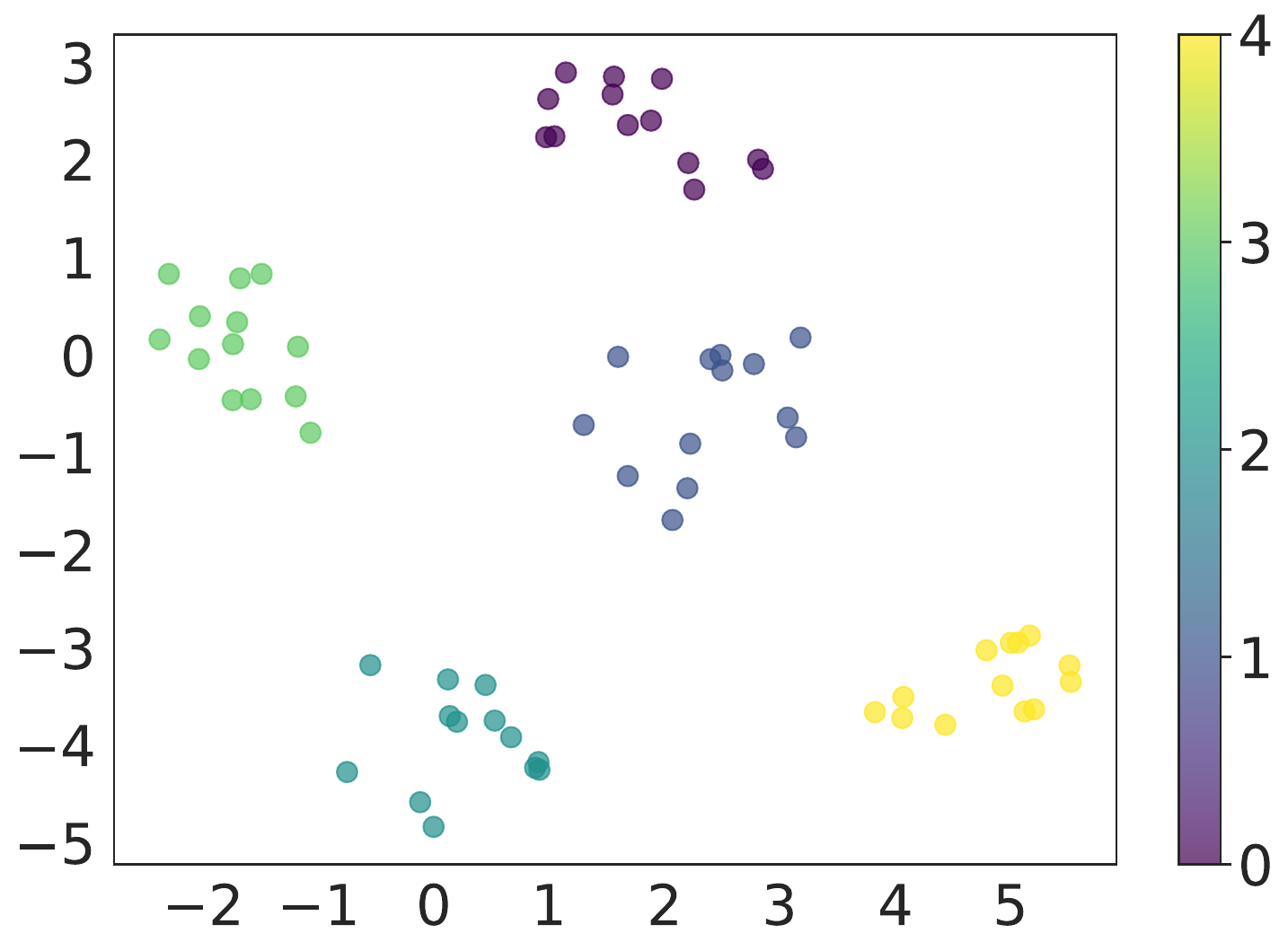}
      }
      \subfloat[Prodigy - 50 shots on FB]{
      \includegraphics[width=0.25\linewidth]{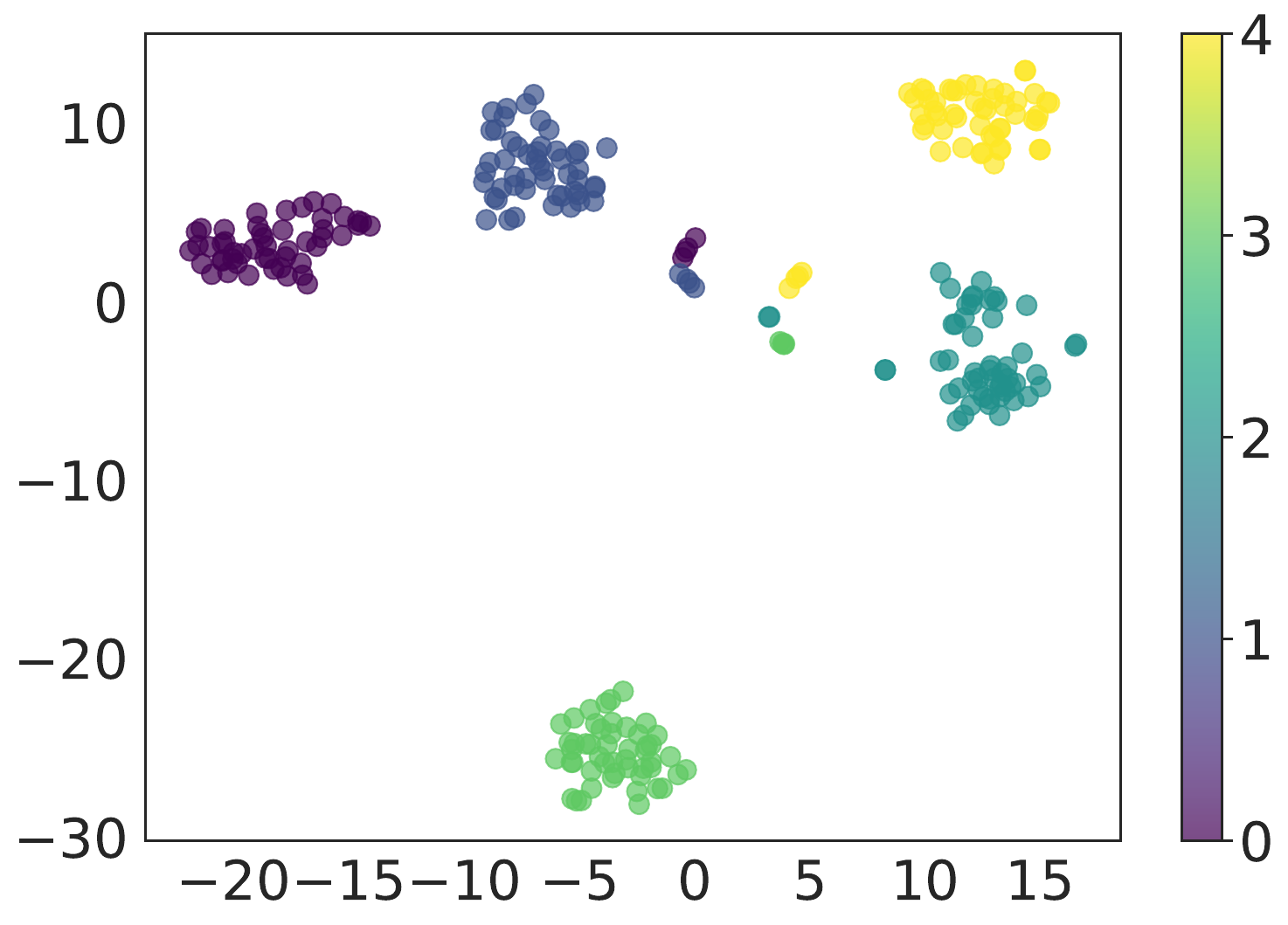}
      }
      \subfloat[GraphPrompter - 50 shots on FB]{
      \includegraphics[width=0.25\linewidth]{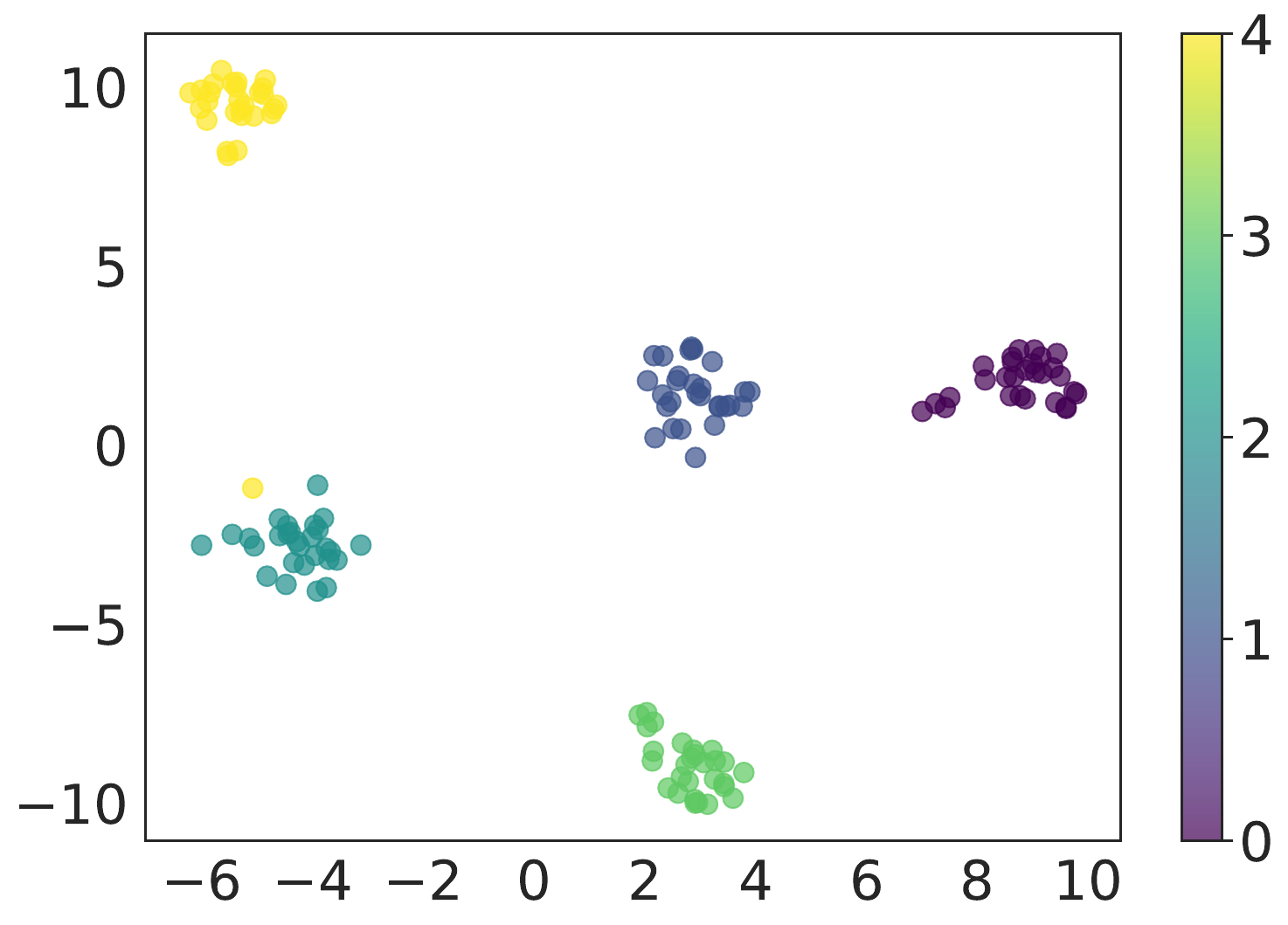}
      }}
      \caption{{Distribution of data node embeddings}, including prompts and query, with different numbers of shots by using Prodigy and our GraphPrompter, on NELL and FB15K-237 (notes "FB") datasets with 5 ways setting. The data node embeddings denote the subgraph embeddings of sampled subgraphs, which are the output of "Data graphs" in Figure~\ref{fig:Prodigy} and the input of "Task graphs". The different colors noted different labels about each data node.}
      \label{fig:diff-prompts-tsne}
\end{figure*}

\subsubsection{\textbf{Enhanced Generalization Ability in Multi-Class Tasks}}
In Sections~\ref{intro} and~\ref{online-aug}, we discussed how the in-context learning capability of the model is influenced by both the quality and diversity of the graph dataset used during pretraining. Due to limitations in the availability of comprehensive datasets and the inherent constraints of the models, it is likely that the model has not been exposed to the full range of possible graph structures during pretraining. This limitation can result in challenges when predicting labels for test datasets that involve a large number of diverse classes.
To assess the robustness of both the Prompt Token and Prompt Graph methods, we evaluate three models—ProG~\cite{allinone}, Prodigy~\cite{huang2023Prodigy}, and our proposed method—across a broader spectrum of class categories. {The results, presented in Table~\ref{tab:more-ways}, focus on the FB15K-237 and NELL datasets, which contain 50, 60, 80 and 100 classes, respectively. As the number of categories increases, we observe a noticeable decline in the model's predictive performance, indicating insufficient generalization by the graph pretraining model when handling multiple classes. 
However, when applying our multistage prompt optimization method, the model's learning ability in context improves approximately 8\%, underscoring the simplicity and effectiveness of our approach in improving the generalization capacity of the model in complex multiclass tasks.}

\subsection{Ablation study}
To verify the effectiveness of our methods, we conducted ablation studies on the FB15k-237 and NELL datasets. The ``w/o $\cdot$" note that our method without a specific component, which is introduced in Section~\ref{method}. The results of these studies are presented in Figure~\ref{fig:ablation-study}. The experimental findings demonstrate the efficacy of our approach in the reconstruction of prompt graphs in the Prompt Generator stage. Additionally, we evaluate the effects of removing $k$NN retrieval and the selection layer separately. The experimental results indicate that, as we discussed in Section~\ref{intro}, the methods of selecting prompts using either $k$NN alone or the pre-trained selection layer alone are not perfect. When both are used simultaneously, they can have a complementary effect. Furthermore, the Prompt Augmentor not only improves predictive performance when there are many categories(e.g., 40 ways) in the downstream data, but it is also effective when there are few categories(e.g., 5 ways). This demonstrates the effectiveness of the method we proposed. Overall, by combining these methods, we achieve state-of-the-art results. {It’s worth noting that the "w/o kNN" model only performs 1\% better than the baseline. While kNN does add some time cost, the performance boost it brings is quite valuable. We can manage the time impact of kNN by adjusting the candidate set size to find a balance between speed and performance.}

\subsection{Hyperparameter analysis}
\label{sec:parameter}

{\subsubsection{\textbf{Analysis of cache size and quality}}
In the Prompt Augmentor stage, we tested different cache sizes, as shown in Figure~\ref{fig:cache}. The results reveal that when the cache size exceeds 3, model performance declines to varying extents, indicating that the noise introduced by additional pseudo-label samples outweighs their benefits. Therefore, we chose a cache size of 3 for our experiments. 
Additionally, we explored the impact of pseudo-label quality on GraphPrompter performance. In the main experiment, we used the highest-confidence samples as pseudo-labels in the cache. In this section, we randomly selected pseudo-labels to test the robustness of our method. Table~\ref{tab:cache} shows the results of random pseudo-label selection, with experiments conducted on the FB15K-237 and NELL datasets using different random seeds and a setting of 20 ways. The results indicate a $2\%$ performance drop when using random pseudo-labels compared to the highest-confidence ones, but the performance still exceeds the baseline. This demonstrates that our pseudo-label selection method is effective and robust to variations in the pseudo-labels.}

\subsubsection{\textbf{Comparison with different GNN architectures}}
We use the GraphSAGE\cite{hamilton2017inductive} as the $\text{GNN}_D$ in Eq~\ref{gi}. The reconstruction process is a kind of graph structures learning, as is GAT\cite{gat2018}, which emphasizes the importance of edges by learning their weights. We complement our experiments using GAT as a prompt generator, shown in Figure~\ref{fig:gat}. As a result, our generator performance is better than GAT, we suppose that the GraphSAGE can scale better in large pre-training graphs.

\subsubsection{\textbf{Evaluation using different numbers of in-context examples}}
To evaluate the contextual learning capability of our method, we conducted performance analyses across varying numbers of prompt examples. The results are presented in Figure~\ref{fig:diff-prompts}, which observed that as the number of prompts increases, both Prodigy and our proposed method initially improve and then diminish performance.  Especially for the arXiv dataset, when the number of prompts exceeds 10, the performance of Prodigy significantly declines. This can be understood as more prompt graphs introducing more noise, thereby affecting the predictive performance of the model. We consider the reason is that the distribution of the pre-training dataset and test dataset are very different when the number of prompts is too large, it is difficult for the pre-trained task graph to aggregate them, and thus the classification performance of the model deteriorates. 

Nonetheless, our method consistently outperforms Prodigy, even with an equivalent number of prompts, thus substantiating the effectiveness of our proposed approach. In Figure~\ref{fig:diff-prompts-tsne}, we display the distribution of data node embeddings by using the t-SNE~\cite{arora2018analysis} algorithm with the different number of prompts. We can see that with the same number of prompts, e.g.shots=50, our GraphPrompter get tighter classification results than the baseline Prodigy, which makes it easier to aggregate the task graphs to get better prediction.

\begin{figure}[t]

  \centering
  \scalebox{0.95}{\includegraphics[width=1\linewidth]{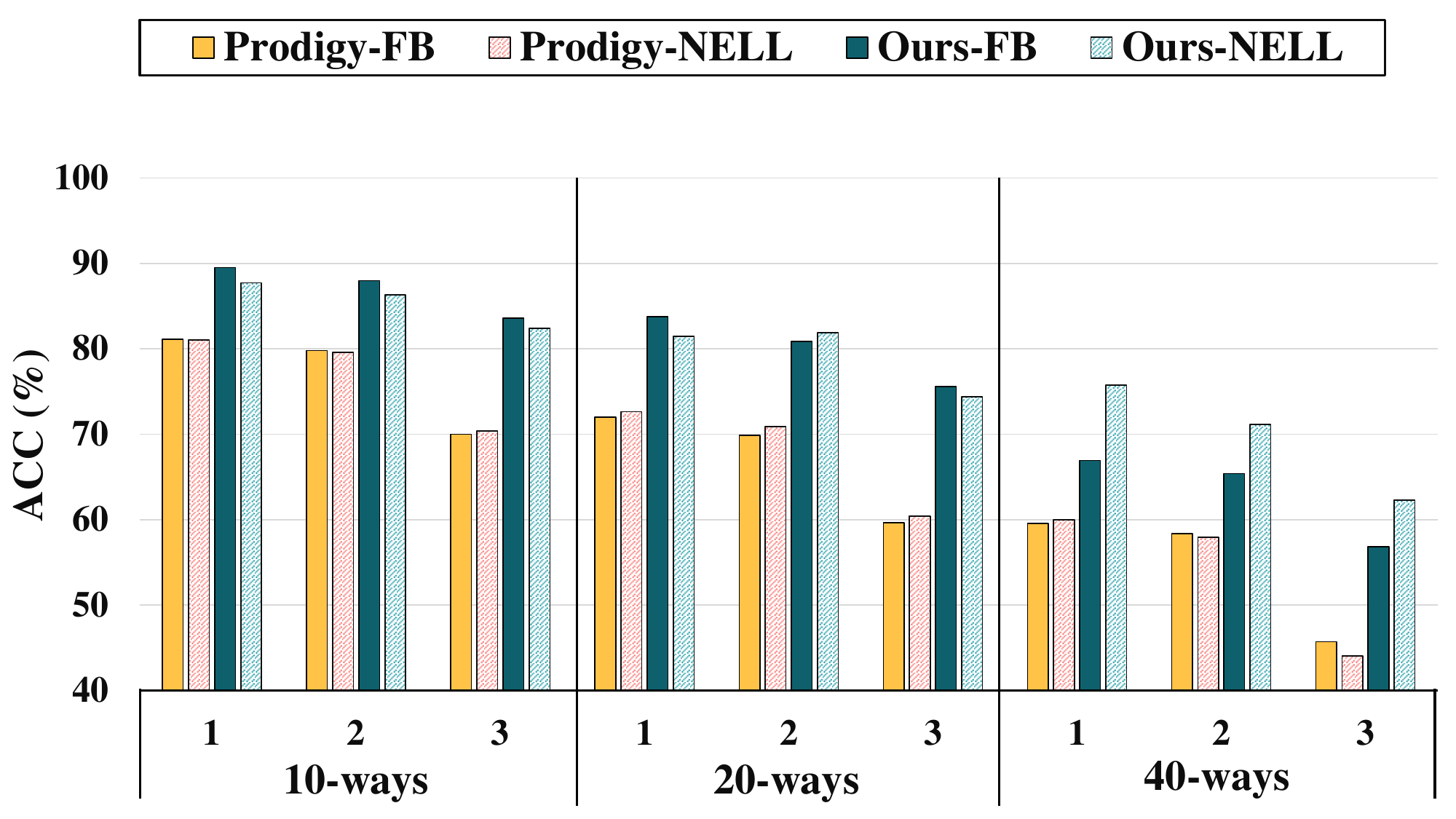}}

  \caption{{Experiment with multi-hop (1/2/3-hop) subgraph on FB15k-237 (note as ``FB") and NELL dataset.}}

  \label{fig:multi-hop}

\end{figure}

{\subsection{\textbf{Multi-hop Analyis}}
To evaluate the performance of our method on multi-hop problems, we conducted experiments on 1, 2, and 3-hop subgraphs, shown as Figure~\ref{fig:multi-hop}, comparing our approach with the baseline. The results show that as the subgraph size increases (i.e., as the logical chain becomes longer), model performance declines. This may be due to the limitations of GNNs, where larger graphs make it more challenging to capture relationships between nodes. Nevertheless, our method still outperforms the baseline, demonstrating its effectiveness on multi-hop subgraphs.}

\subsection{\textbf{Time Efficiency Analysis}} \label{test-time}

\begin{figure}
      \centering 
      \scalebox{0.95}{
      \subfloat[Training Loss]{
      \includegraphics[width=0.5\linewidth]{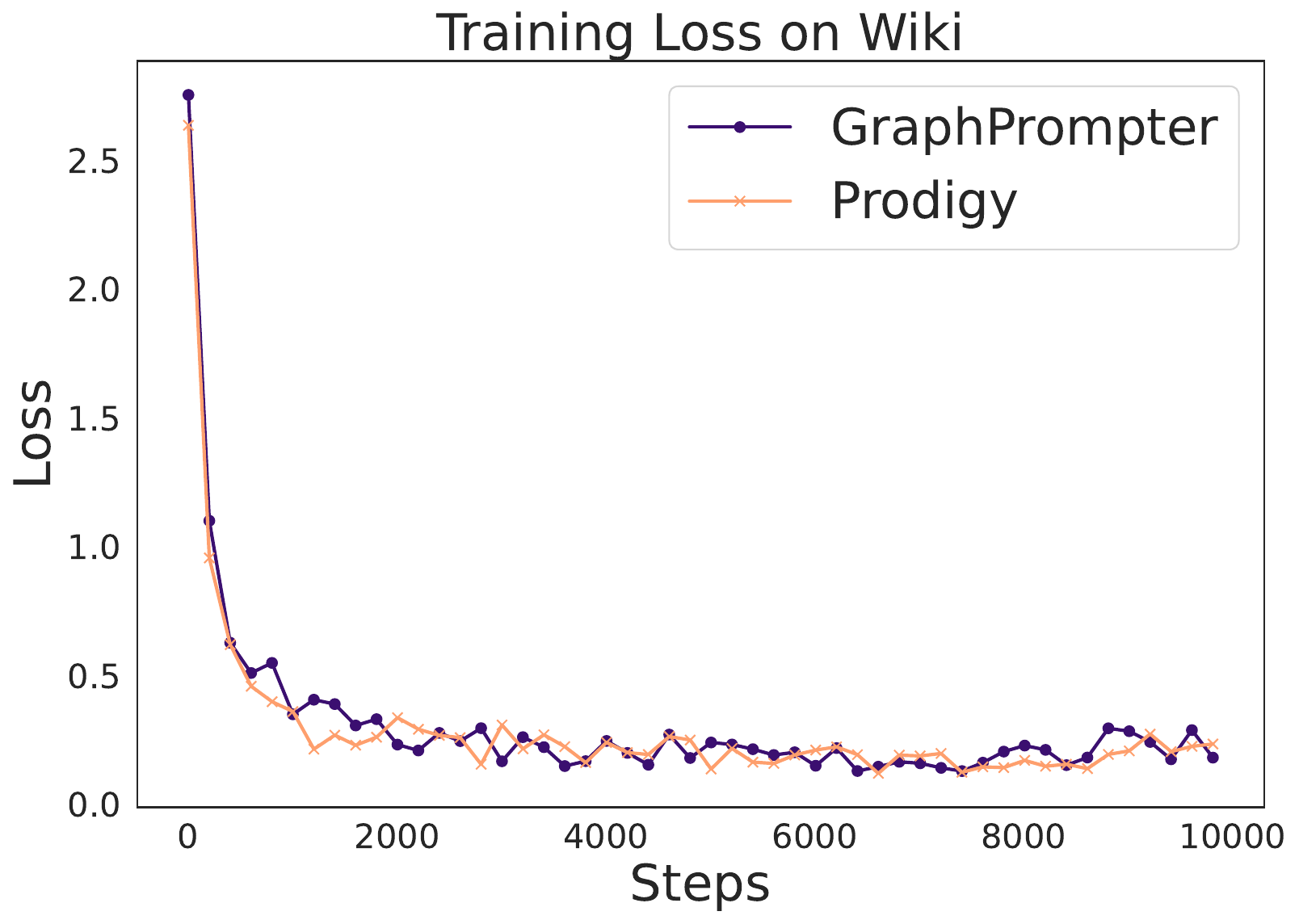}}
    
      \subfloat[Training Accuracy]{
      \includegraphics[width=0.5\linewidth]{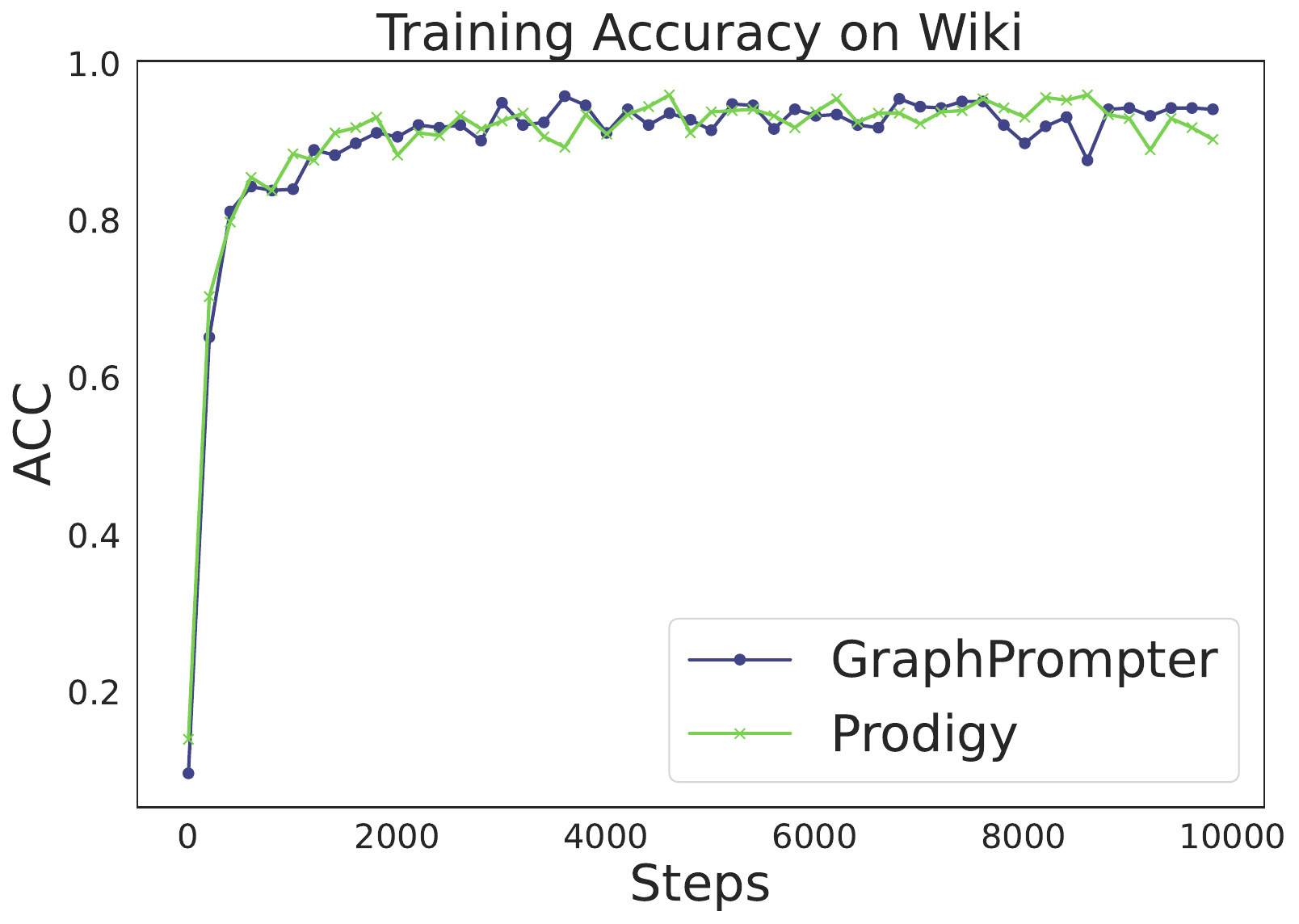}
      }}
      
    \caption{\textcolor{black}{The Comparison of Training Loss and Training Accuracy over 10k steps on Wiki dataset.}}
      \label{fig:loss-acc}
    
  \end{figure}

We present the training details of our model in Figure~\ref{fig:loss-acc} and compare it with Prodigy, including the loss convergence curve and the increase in training accuracy. As shown in the figure, our model performs comparably to the baseline in terms of both convergence speed and accuracy. Both the reconstruction layers and selection layers do not incur significant computational costs during pretraining, as we use a two-layer neural network and the same pretraining steps as Prodigy. The additional computational complexity introduced by the MLP is negligible compared to the overall cost of the GNNs. All of our experiments were conducted using a single NVIDIA A100 (40GB) GPU. A single pretraining run of 10k steps typically takes approximately 3 hours, which is consistent with Prodigy's performance.

{Our inference time overhead primarily stems from prompt retrieval in the selector and augmentator in Task graphs. For $N$ candidates, $k$ prompts, $m$ ways, and $d$ dimensions: 
\begin{equation}
    T_{\text{Prodigy}}=O((k+q)md^2), 
\end{equation}
\begin{equation}
    T_{\text{Ours}}=O((N+q)md) + O((2k+q)md^2).
\end{equation}
Due to the incorporation of an additional $c=3$ pseudo-label prompts via the cache mechanism, where $\hat{S}=\hat{S}\cup C$ and $|\hat{S'}|=2*k=6$, the inference time of the model has increased by a factor of two to three. The retrieval module we use is pluggable. We can balance model performance and speed based on the specific scenario. In real-world applications, we can remove the retrieval module or reduce the size of the candidate set to find the optimal solution. } The results, shown in Table~\ref{tab:time}, indicate that while GraphPrompter adds some additional time for prompt selection, it does not introduce a significant computational burden.

\begin{table}[t]
  \caption{\textcolor{black}{The inference time of per query on FB15K-237 and NELL dataset with 10, 20, and 40 classes respectively.}}
  \label{tab:time}
  \scalebox{0.96}{\begin{tabular}{c|ccccccc}
    \toprule
     & \multicolumn{3}{c}{FB15K-237} & \multicolumn{3}{c}{NELL} \\
    Classes  & 10 & 20 & 40 & 10 & 20 & 40 \\
    \midrule
    Prodigy  & 34ms & 68ms & 106ms & 26ms & 42ms & 82ms \\
    \midrule
    GraphPrompter & 90ms & 150ms  & 280ms & 80ms & 120ms & 240ms \\
    \bottomrule
  \end{tabular}} 

\end{table}

\subsection{Prompt Examples}
\begin{figure}[t]
  \centering
  \scalebox{1}{\includegraphics[width=1\linewidth]{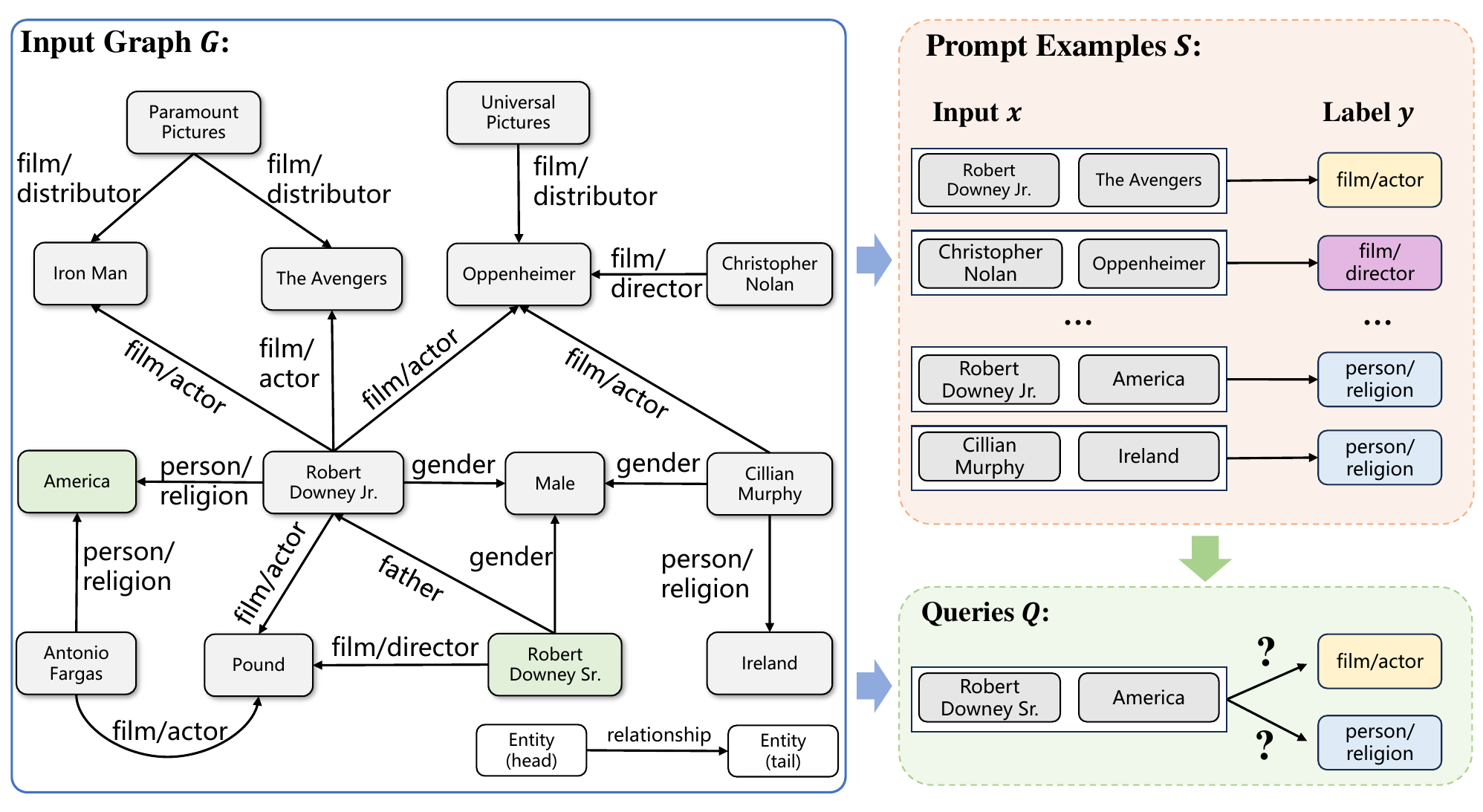}}
  \caption{{A prompt example of Graph In-Context Learning.}}
  \label{fig:casestudy}
\end{figure}

Figure~\ref{fig:casestudy} shows a specific prompt example to illustrate the process of Graph In-Context Learning. First, the input graph \( G \) is a knowledge graph containing multi-hop logical relationships. Our goal is to predict the edge category of a query \( q \in Q \), where \( Q \subset G \), represented in the graph as the relationship between "Robert Downey Sr." and "America." We then randomly sample some triples from \( G \) and extract subgraphs, where their embeddings are denoted as \( x \) and edge categories as \( y \), forming a bipartite Task Graph, as described in Section~\ref{method}. Finally, our algorithm infers the answer to the query based on the examples provided in the prompt. It is important to note that the main contribution of GraphPrompter lies in optimizing the prompt. Specifically, we optimize the generation of embeddings, the selection, and supplementation of prompts. For example, our method tends to select "Robert Downey Jr., America" as the prompt rather than "Robert Downey Jr., The Avengers," since the former is closer to the query in the latent space.
\section{Conclusion}
To tackle the limitations of prompt construction strategies of existing graph in-context learning methods and enhance their generalization ability, we have for the first time validated the impact of prompt quality on the performance of the graph foundation model based on prompt graphs, and we proposed a novel multi-stage adaptive prompt optimization method GraphPrompter, including Prompt Generator, Prompt Selector, and Prompt Augmentor.
With Prompt Generator and Prompt Selector, we can generate and select the most suitable prompts for each query based on graph topology and semantic similarity, thereby significantly improving the predictive accuracy of the model. Furthermore, we also propose Prompt Augemntor, a non-parametric adaptive method for testing, based on the LFU cache replacement strategy, to enhance the generalization performance of the pre-trained graph model on new datasets. Our GraphPrompter achieves superior generalization to downstream graphs without gradient updates, consistently outperforming all baselines on node classification and link prediction tasks.

\textbf{Futher Discussion:} We have for the first time validated the impact of prompt quality on the performance of the graph foundation model based on prompt graphs, and proposed improved solutions from the generation, selection, and use stages of prompts. The framework we proposed has excellent applicability and scalability. Specifically, the reconstruction layer in the prompt generator can be replaced with networks other than just MLP. In the retrieval stage, we can also use other clustering methods to dynamically and adaptively select prompts. Similarly, we can replace the cache in the prompt augmenter with other caching solutions. Extensive results reflect the advantages of the method we proposed and demonstrate the potential of graph in-context learning.


\section{Acknowledgment}

This research was supported by grants from the National Key Research and Development Program of China (Grant No. 2024YFC3308200), the Key Technologies R \& D Program of Anhui Province (No. 202423k09020039), and the Fundamental Research Funds for the Central Universities.

\bibliographystyle{ieeetr}
\bibliography{sample-base}

\end{document}